\documentclass{article}
\PassOptionsToPackage{table}{xcolor}
\usepackage{microtype}
\usepackage{graphicx}
\usepackage{subcaption}
\usepackage{booktabs}
\usepackage{xurl}
\usepackage{hyperref}

\usepackage[preprint]{icml2026}
\usepackage{amsmath,amssymb,mathtools,amsthm}
\usepackage[capitalize,noabbrev]{cleveref}
\usepackage{array}
\usepackage{multirow}
\usepackage{tabularx}
\newcolumntype{C}{>{\centering\arraybackslash}X}
\usepackage{placeins}
\usepackage{flushend}

\theoremstyle{plain}
\newtheorem{theorem}{Theorem}[section]
\newtheorem{proposition}[theorem]{Proposition}

\theoremstyle{definition}

\crefname{assumption}{Assumption}{Assumptions}
\Crefname{assumption}{Assumption}{Assumptions}

\definecolor{motagray}{gray}{0.94}
\definecolor{motaink}{HTML}{253952}
\definecolor{referencegray}{gray}{0.975}
\definecolor{referenceink}{gray}{0.50}
\definecolor{gainred}{HTML}{B44242}
\definecolor{lossblue}{HTML}{326BB0}
\newcommand{\gain}[1]{\textcolor{gainred}{(#1)}}
\newcommand{\loss}[1]{\textcolor{lossblue}{(#1)}}
\newcommand{\zerodelta}[1]{\textcolor{referenceink}{(#1)}}
\newcommand{\scorecell}[2]{\mbox{#1\,\textnormal{\scriptsize #2}}}
\newcommand{\mota}{\textup{\textsc{MoTA}}}
\newcommand{\E}{\mathbb{E}}
\newcommand{\R}{\mathbb{R}}
\newcommand{\KL}{\operatorname{KL}}
\newcommand{\Cov}{\operatorname{Cov}}
\newcommand{\Var}{\operatorname{Var}}
\newcommand{\softmax}{\operatorname{softmax}}
\newcommand{\RMSNorm}{\operatorname{RMSNorm}}

\newcommand{\spn}{\operatorname{span}}
\newcommand{\norm}[1]{\left\lVert #1\right\rVert}
\newcommand{\oursrow}{\rowcolor{motagray}}
\newcommand{\headcell}[2][c]{\shortstack[#1]{#2}}
\graphicspath{{figures/}}
\icmltitlerunning{MoTA: Moment-Tilted Attention for Looped Transformers}

\begin{document}
\raggedbottom
\twocolumn[
\icmltitle{MoTA: Moment-Tilted Attention for Looped Transformers}

\begin{icmlauthorlist}
\icmlauthor{Long Qian}{casia,ucas}
\icmlauthor{Bingke Zhu}{casia,ucas,wuhan}
\icmlauthor{Yu Li}{casia,ucas}
\icmlauthor{Xiaomian Kang}{casia,ucas}
\icmlauthor{Jiajun Zhang}{casia,ucas,wuhan}
\icmlauthor{Yingying Chen}{casia,ucas,wuhan}
\icmlauthor{Jinqiao Wang}{casia,ucas,wuhan}
\end{icmlauthorlist}

\icmlaffiliation{casia}{
Foundation Model Research Center,
Institute of Automation,
Chinese Academy of Sciences,
Beijing, China
}

\icmlaffiliation{ucas}{
University of Chinese Academy of Sciences,
Beijing, China
}

% \icmlaffiliation{ucas1}{
% School of Future Technology,
% University of Chinese Academy of Sciences,
% Beijing, China
% }

% \icmlaffiliation{ucas2}{
% School of Advanced Interdisciplinary Sciences,
% University of Chinese Academy of Sciences,
% Beijing, China
% }

\icmlaffiliation{wuhan}{
Wuhan AI Research,
Wuhan, China
}

\icmlcorrespondingauthor{Bingke Zhu}{bingke.zhu@nlpr.ia.ac.cn}
% \icmlkeywords{looped transformers, attention, mathematical reasoning, language models}
\vskip 0.3in
]
\printAffiliationsAndNotice{}

\begin{abstract}
\setlength{\parfillskip}{0pt plus .35\linewidth}
Looped Transformers increase the computational depth of language models by repeatedly applying shared layers. However, standard looped Transformers rely on hidden-state updates to guide subsequent attention reads, leaving the next retrieval goal implicit. In this paper, we introduce Moment-Tilted Attention (MoTA), which uses the same attention weights to retrieve content and write a compact target for the next loop. To guide the next read, MoTA uses soft moment matching to move the attention-weighted mean of learned source features toward the target while preserving proximity to the original attention distribution. The resulting update uses source covariance to determine the correction direction and the range of projected source scores to set a closed-form step size. We further establish a descent guarantee for the dual objective. Experiments across four model sizes demonstrate consistent improvements in aggregate pretraining performance over standard looped Transformers. The gains extend to mathematical knowledge after lightweight adaptation, with further improvements in mathematical problem solving, scientific question answering, and analytical reasoning at larger model scales after mathematical midtraining.

% Looped Transformers increase a language model's effective depth by repeatedly applying shared layers. However, in standard looped Transformers, each attention read updates the hidden state without explicitly specifying what the next read should retrieve.
% This suggests that each attention read should turn the information it retrieves into a target for the next read, interpreted in the context of the sources available at that step. Guided by this insight, we introduce Moment-Tilted Attention (\mota), in which each attention read writes a compact learned target for the following loop. A moment-matching objective that encourages the attention-weighted mean of learned source features toward this target yields a covariance-based correction and a closed-form step size, allowing a single attention distribution to retrieve content and write the next target.
% Scaling experiments from 140M to 1.3B parameters demonstrate strong improvements in general pretraining, with further gains in mathematical adaptation, scientific question answering, and analytical reasoning following mathematical midtraining.
% Theoretical analysis characterizes how source covariance and score range shape the attention update, while architectural ablations examine the contributions of its components and placement.

\end{abstract}
\setlength{\parfillskip}{0pt plus .35\linewidth}
\section{Introduction}
\label{sec:introduction}

Looped Transformers increase a language model's computational depth by repeatedly applying shared layers~\cite{dehghani2019universal,geiping2025recurrent}.
Parameter sharing allows additional computation without a proportional increase in model size.
Recent studies have demonstrated the reasoning potential of recurrent depth and improved training stability~\cite{saunshi2025latent,prairie2026parcae}.
Each loop can refine the current representation, but useful refinement also depends on the information selected for the update.
An important architectural question therefore remains: \emph{how should information retrieved in one loop guide attention in the next?}

\begin{figure}[t]
  \centering
  \includegraphics[width=\linewidth]{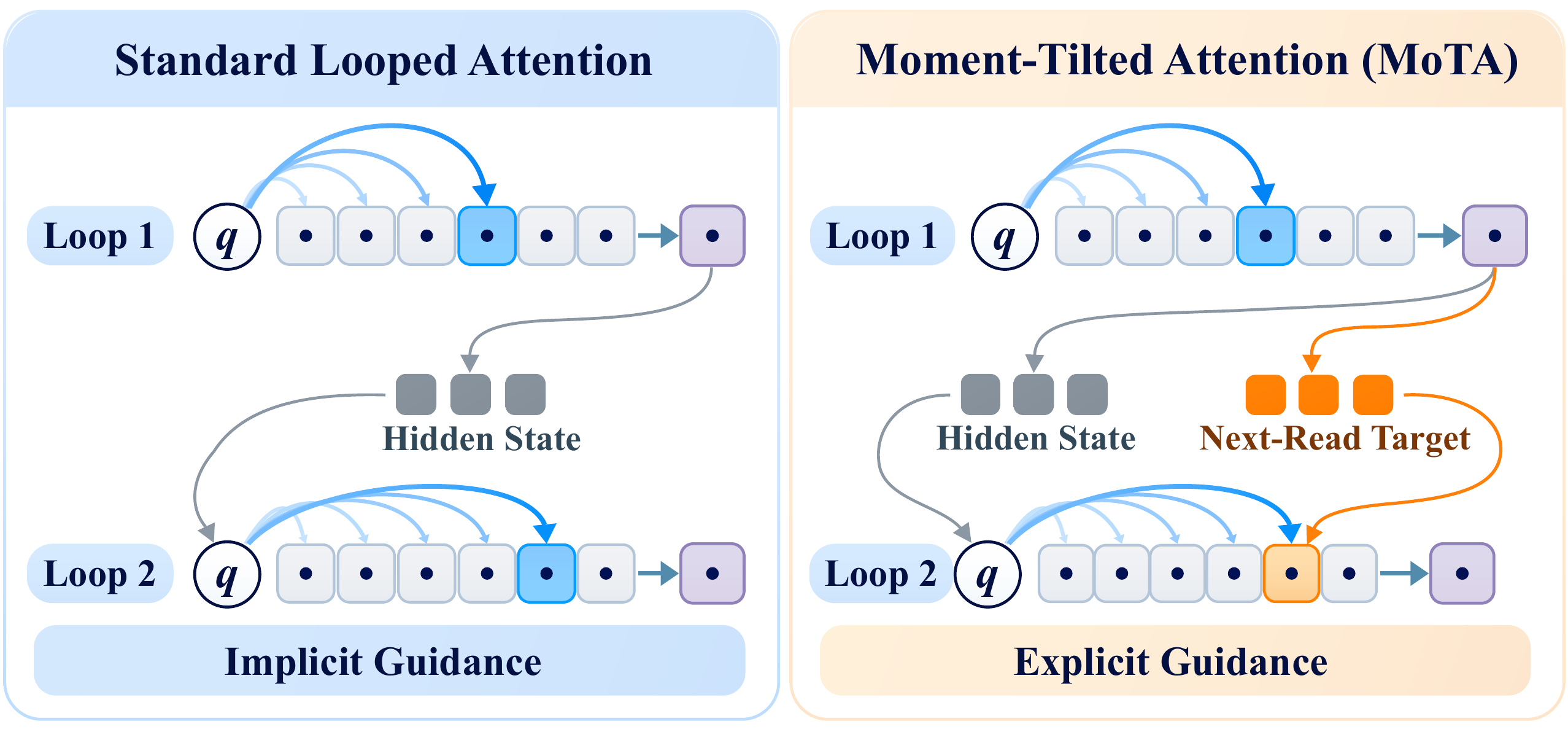}
  \caption{Standard looped attention guides subsequent reads through hidden-state updates. MoTA extends the recurrence with an explicit guidance, where each read retrieves content and writes a compact target that directly guides source selection in the next loop.}
  \label{fig:motivation}
\end{figure}

In standard looped Transformers, attention assigns weights to the current and preceding token representations and aggregates their values~\cite{vaswani2017attention}.
The resulting readout updates the hidden state, and the updated state produces the queries, keys, and values for the next loop.
Information retrieved in one loop can change which sources matter in the next.
Consequently, the hidden state carries both retrieved content and information used for subsequent selection.
However, the next retrieval objective remains implicit in the hidden-state update.
The readout summarizes the outcome of the current selection, whereas a retrieval target specifies the desired outcome of a later read.
Explicitly representing the target would therefore provide a direct link between successive attention reads.

In this paper, we introduce \emph{Moment-Tilted Attention} (MoTA) to make the connection between successive reads explicit (Figure~\ref{fig:motivation}).
Specifically, MoTA uses the same attention weights to retrieve content and write a compact target for the next loop.
Each source value is mapped to separate learned read and write features.
The weighted average of the write features becomes a target for the next loop's average read features.
In the next loop, MoTA compares the target with the mean read features under query-key attention and uses the discrepancy to correct the attention weights.
The read and write projections are learned jointly with the language model through next-token prediction.
During recurrence, each token and attention head carries a separate target between visits to the same shared layer.
The main configuration stores two numbers per target and applies MoTA in the first attention layer of the shared core.

To turn the learned target into an attention update, MoTA must account for the available sources.
Reweighting can change a feature average only along directions in which the source features differ.
MoTA therefore uses source statistics to determine how attention should change.
We formulate the update through soft moment matching.
The objective encourages the attention-weighted feature mean to approach the target.
A Kullback-Leibler penalty simultaneously controls deviation from the base distribution.
MoTA uses source covariance to convert the target discrepancy into a correction direction.
The covariance accounts for variation and correlations among the features.
Along the resulting direction, MoTA takes one analytic step, with a closed-form step size set by the range of source score corrections.
We establish a descent guarantee for the associated dual objective.
Finally, the corrected weights retrieve the original values and write the next target, completing the connection between successive reads.

Experimentally, we evaluate MoTA on four Parcae model families, from 140M to 1.3B parameters, through general pretraining, lightweight math adaptation, and further math midtraining.
General pretraining tests broad language capabilities.
The mathematics stages then examine adaptation with a small text budget and further learning with a larger cumulative budget.
Using the same pretraining token budgets, MoTA improves CORE and CORE-Extended, Parcae's aggregate evaluation scores, over the corresponding looped baselines at every model scale.
At 1.3B, the gains reach 3.56 and 2.01 points, respectively.
The benefits also extend to specialized reasoning.
Following lightweight adaptation, MoTA outperforms the looped baselines on MMLU College Mathematics and Abstract Algebra at 770M and 1.3B.
After further math midtraining, the 1.3B model exceeds the looped baseline by 1.40 percentage points on MATH and 3.11 on GSM8K.
Beyond mathematics, the 1.3B model improves scientific question answering and analytical reasoning.
Structural ablations further show the value of full source covariance for mathematical problem solving.
MoTA thus provides a direct way to learn information selection across loops, with benefits in both general language learning and specialized reasoning.

In summary, our contributions are threefold:
\begin{itemize}
  \item We introduce MoTA, a recurrent attention mechanism in which each read retrieves content and writes a learned target for subsequent information selection.
  \item A soft moment-matching formulation yields an analytic attention update governed by source covariance and score differences, with a descent guarantee for the associated dual objective.
  \item Experiments across four model scales and three evaluation stages demonstrate gains in general language learning and specialized reasoning. Structural ablations examine the roles of target feedback, source covariance, and placement within the shared core.
\end{itemize}

\section{Related Work}
\label{sec:related}

\paragraph{Looped Transformers and reasoning.}
Universal Transformers introduce recurrence over depth
\cite{dehghani2019universal}, ALBERT shares parameters across layers
\cite{lan2020albert}, and deep equilibrium models express weight-tied
computation through a fixed point, a hidden state left unchanged by the
shared transformation \cite{bai2019equilibrium}.
Recurrent-depth language models repeatedly update hidden states to
improve reasoning \cite{geiping2025recurrent}; Parcae studies how this
computation scales with model size and training data
\cite{prairie2026parcae}. These architectures connect to a broader
algorithmic view: recurrent networks can solve harder instances by
taking additional steps \cite{schwarzschild2021algorithm}, and looped
Transformers can learn iterative algorithms from examples supplied in
their input \cite{yang2024looped}.  Studies of hidden-state recurrence, iteration encoding, and gradient descent implemented through attention further reveal the computational capabilities of shared layers~\cite{saunshi2025latent,xu2025expressive,vonoswald2023gradient}.
For mathematical reasoning, chain-of-thought prompting exposes
intermediate steps in generated text \cite{wei2022chain}, while Minerva
and Llemma develop quantitative and mathematical capabilities through
specialized training \cite{lewkowycz2022minerva,azerbayev2024llemma}.
\mota\ addresses information selection across recurrent iterations by letting each attention read write a target that directly conditions the next selection. Our three evaluation stages examine how this mechanism
learns general representations and uses them for mathematical tasks.

\paragraph{Attention and adaptive memory.}
Transformer attention provides content-dependent retrieval
\cite{vaswani2017attention}, and modern Hopfield networks connect this
operation to associative memory, which retrieves stored patterns by
minimizing an energy function over its state
\cite{ramsauer2021hopfield}. Feed-forward layers encode retrievable
key-value associations, linking stored knowledge to its use during
a forward pass \cite{geva2021memories}. Linear attention admits both a
recurrent formulation \cite{katharopoulos2020linear} and a fast-weight
interpretation, in which a memory matrix is updated as tokens arrive
\cite{schlag2021fastweights}. Gated Delta Networks and
test-time training layers develop learned memory updates
\cite{yang2025gateddelta,sun2025ttt}, while Attention Residuals select
among earlier depth representations \cite{kimi2026attentionresiduals}.
\mota\ retains the causal token sources and their content values.
Its additional state specifies a desired mean in learned source
coordinates, and the current source statistics determine how that
state changes attention. This gives the recurrent state a direct role
in selecting the information used by the next loop.

\paragraph{Attention distributions and geometry.}
Entmax constructs sparse attention distributions through an alternative
regularizer \cite{peters2019entmax}. \mota\ instead uses a KL penalty
relative to the base attention together with a target for learned
source moments. Exponentially reweighting the base distribution yields
an objective whose curvature is determined by source covariance.
This connects the update to natural-gradient methods, which account for
how parameter changes alter a probability distribution
\cite{amari1998natural,martens2020natural}.
The covariance translates a target discrepancy into a direction of
score change; the directional range across the current sources then
sets an analytic step. Our derivation connects these two quantities to
invariance, descent, and the response of the retrieved content.
The same attention distribution reads that content and writes the next
target, closing the mathematical construction within the shared core.
\Cref{app:comparison} develops the architectural and mathematical
comparisons, including residual attention, head composition,
differential attention, and online neural memory.

\section{Preliminary}
\label{sec:preliminary}

\paragraph{Attention as a distribution over sources.}
For one head at query position $i$, let $q_i,k_j,v_j\in\R^{d_h}$ denote
the query, key, and value, each of width $d_h$, after the backbone's
position and normalization operations. Causal attention reads only the
current and preceding token positions, called sources, $j\leq i$:
\begin{equation}
 \begin{aligned}
 \ell_{ij}&=\frac{q_i^\top k_j}{\sqrt{d_h}},\qquad
 p_{ij}=\softmax_{j\leq i}(\ell_{ij}),\\
 y_i&=\sum_{j\leq i}p_{ij}v_j.
 \end{aligned}
 \label{eq:attention}
\end{equation}
Here $\ell_{ij}$ is the unnormalized attention score, or logit, and $p_i$
is the base attention distribution before any \mota\ correction.
We write $\E_a[x]=\sum_j a_jx_j$ for an attention-weighted mean and
$\norm{x}$ for the Euclidean norm. Equivalently, $p_i$ maximizes the
expected score plus entropy over all probability vectors on the sources.
This variational interpretation provides a natural starting point for
introducing a reading goal while retaining the original relevance scores.

\paragraph{Recurrence over depth.}
Following the prelude-core-coda organization of recurrent-depth models
\cite{geiping2025recurrent}, a prelude stack of Transformer layers
produces the input representation $E$ before recurrence. Let $H^t$
denote the hidden state after loop $t$, with $H^0$ denoting the
initial recurrent state. A shared Transformer core
$\mathcal R_\theta$, parameterized by weights $\theta$ reused
across iterations, updates the state as
\begin{equation}
    H^t = \mathcal R_\theta(DH^{t-1}+UE).
    \label{eq:loop}
\end{equation}
Here, $D$ is a positive diagonal matrix with entries below one
that scales the previous hidden state, while $U$ is a learned
transformation applied to the input representation $E$.
The term $UE$ injects the input at every iteration, following
the Parcae backbone~\cite{prairie2026parcae}.
After the final loop, a non-recurrent stack of Transformer layers,
called the \emph{coda}, processes the final state before the output
head predicts the next token. We use \emph{visit} to denote one
execution of a shared layer during a loop. In this architecture,
$H^{t-1}$ influences the next attention read through its queries,
keys, and values. \mota\ augments the recurrent state with a
target $m_i^t\in\R^r$, written at token position $i$ during
loop $t$ to guide the next visit to the same attention layer.
Each attention head maintains its own target, with the head index
omitted for clarity. We use $r=2$ in the main models and analyze
general $r$ throughout the derivation.

\paragraph{Learned source coordinates.}
For a given attention read, we omit token and iteration indices
and denote the incoming target by $m$. This target specifies a
desired attention-weighted mean of learned source features.
Each source $j$ has two feature vectors $n_j,w_j\in\R^r$,
obtained through separate learned projections of its normalized
value vector $v_j$. We call these low-dimensional features
\emph{source coordinates}. The read coordinates $n_j$ describe
properties used to select sources, while the write coordinates
$w_j$ describe properties to be requested in the next iteration.
Under the same attention distribution, reading aggregates the
original values into content, while writing aggregates the
write coordinates into the next target. The separate projections
allow the selected information to specify a different retrieval
objective for the following iteration. All projections are
learned jointly through language modeling.
\section{Method}
\label{sec:method}

\begin{figure*}[t]
 \centering
 \includegraphics[width=\textwidth]{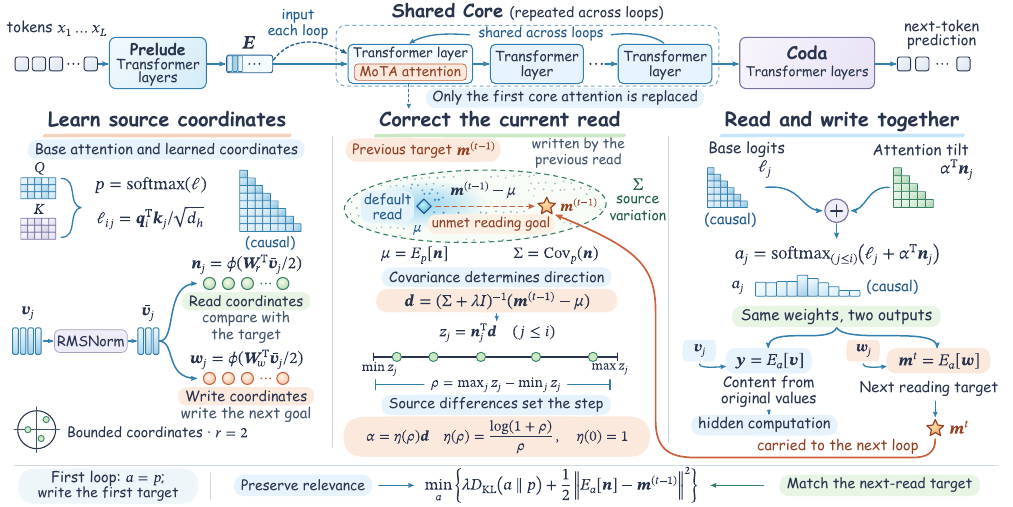}
 \caption{\textbf{Moment-Tilted Attention inside the shared recurrent core.}
 For each token and head, learned read coordinates $n$ describe the sources
 used for target matching, while write coordinates $w$ produce the next
 reading target. The discrepancy between the previous target and the
 default mean $\mu$ under base attention $p$ is translated through source
 covariance $\Sigma$ into a correction direction $d$; the projected
 score range $\rho$ sets the step $\alpha=\eta(\rho)d$.
 Corrected weights $a$ read original content $v$ and write the next
 target. The peach arrow carries that target to the next visit. Only the
 first attention layer of the shared core is replaced.}
 \label{fig:method}
\end{figure*}

\mota\ turns a previous reading result into a target for a new attention
distribution. The construction has one organizing objective: preserve the
base attention's relevance while moving its expected read coordinates
toward that target. \Cref{fig:method} connects this objective to the
read and write operations within the shared recurrent core.

\subsection{A Read That Writes Its Successor}
\label{sec:readwrite}

Each head learns two projections $W_r,W_w\in\R^{d_h\times r}$ from
normalized source values $\bar v_j=\RMSNorm(v_j)$, where root mean square
normalization (RMSNorm) divides each vector by its root mean square:
\begin{equation}
 n_j=\phi\!\left(\tfrac12W_r^\top\bar v_j\right),\qquad
 w_j=\phi\!\left(\tfrac12W_w^\top\bar v_j\right),
 \label{eq:coordinates}
\end{equation}
where $\phi(z)=\tanh(\norm{z})z/\norm{z}$ with $\phi(0)=0$.
This smooth radial map preserves direction and bounds the vector norm
by one, so $\norm{n_j},\norm{w_j}\leq1$.
Given an attention distribution $a_i^t$, the layer returns
\begin{equation}
 y_i^t=\E_{a_i^t}[v],\qquad m_i^t=\E_{a_i^t}[w].
 \label{eq:jointread}
\end{equation}
Both outputs are derived from the same attention distribution but serve different purposes. The content $y_i^t$
enters the usual output projection and residual computation; the target
$m_i^t$ is retained for the next visit to this attention layer. Separate
read and write projections let the model learn what a source means for
current selection and what it should request afterward.

\subsection{From a Reading Target to an Attention Objective}
\label{sec:objective}

For one query, omit $i,t$ and let $m$ be the previous target. We define
the desired reading distribution through
\begin{equation}
 a^\star=\arg\min_{a\in\Delta}
 \left\{\lambda\KL(a\Vert p)
       +\tfrac12\norm{\E_a[n]-m}^{2}\right\},
 \label{eq:primal}
\end{equation}
Here $\Delta=\{a:a_j\geq0,\sum_j a_j=1\}$ is the probability simplex
over the causal sources, and
$\KL(a\Vert p)=\sum_j a_j\log(a_j/p_j)$ is the Kullback-Leibler divergence.
The coefficient $\lambda>0$ balances preserving the base distribution
against matching the target. The KL term keeps the new read near $p$.
The second term asks it to realize the target in the space of source
properties. We call this \emph{soft moment matching}: the first moment,
or weighted mean, is encouraged to approach the target through a penalty.
This gives a well-defined compromise even when the target lies beyond
the convex hull, the set of all weighted means attainable from the
current read coordinates.

The equivalent dual formulation replaces one probability per source
with $r$ score coefficients $\alpha\in\R^r$ and minimizes
\begin{equation}
 G(\alpha)=\log\sum_j p_j e^{\alpha^\top n_j}
       +\frac{\lambda}{2}\norm{\alpha}^{2}-m^\top\alpha,
 \label{eq:dual}
\end{equation}
whose solution induces $a_j^\star\propto p_j
e^{\alpha^{\star\top}n_j}$.
Multiplying $p_j$ by $e^{\alpha^\top n_j}$ and renormalizing is an
\emph{exponential tilt} of the base distribution
\cite{wainwright2008graphical}. The term $\alpha^\top n_j$ is the
additive logit correction labeled \emph{Attention tilt} in
\Cref{fig:method}.
The dual represents a full-prefix attention distribution through $r$
tilt parameters. \mota\ takes one analytic step in this space on each
visit to the shared attention layer.

\subsection{Let the Sources Determine the Correction}
\label{sec:direction}

At $\alpha=0$, the base attention supplies the mean and covariance
that govern the local update:
\begin{equation}
 \mu=\E_p[n],\qquad
 \Sigma=\E_p[nn^\top]-\mu\mu^\top.
 \label{eq:moments}
\end{equation}
At the base distribution, the gradient and Hessian (matrix of second
derivatives) of $G$ are $\mu-m$ and $\Sigma+\lambda I$, respectively,
where $I$ is the $r\times r$ identity matrix. Applying the inverse Hessian to the negative gradient yields the Newton direction
\begin{equation}
 d=(\Sigma+\lambda I)^{-1}(m-\mu).
 \label{eq:direction}
\end{equation}
The residual $m-\mu$ describes the requested change. The covariance
translates that request into a change in source scores, taking account
of the variation and correlation already present in the sources.
The positive addition $\lambda I$, called a ridge, makes the matrix
invertible and limits the correction in directions with little variation.
We set $\lambda=0.1$.

The magnitude of the correction should be determined in the same source
space. Attention responds to score differences, so we measure the range
of the projected scores:
\begin{align}
 \rho&=\max_j n_j^\top d-\min_j n_j^\top d,\label{eq:range}\\
 \alpha&=\eta(\rho)d,\qquad
 \eta(\rho)=\frac{\log(1+\rho)}{\rho},\quad\eta(0)=1.
 \label{eq:step}
\end{align}
The step strength depends on distinctions present in the sources.
\Cref{sec:theory} derives it from the directional curvature of $G$ and
establishes invariance to target shifts along directions in which the
sources agree.

The resulting distribution combines the original relevance scores
with the learned moment correction:
\begin{equation}
 a_{ij}^t=\softmax_{j\leq i}
       \left(\ell_{ij}^t+(\alpha_i^t)^\top n_j^t\right).
 \label{eq:tilt}
\end{equation}
\Cref{eq:jointread} then closes the recurrence. On the first loop, the
target is absent: we set $\alpha=0$, read with $p$, and write the first
target. Subsequent loops apply the update above. A zero vector remains a valid target and is handled by the same update equations.

\subsection{Architecture and Training}
\label{sec:implementation}

We replace the first attention layer of the shared core with \mota.
The remaining attention layers, multilayer perceptrons (MLPs), residual
paths, prelude, coda, and input injection follow the backbone.
For $H$ heads and hidden width $d_{\mathrm{model}}=Hd_h$, the two
headwise projections add $2d_{\mathrm{model}}r$ parameters, or
$4d_{\mathrm{model}}$ at $r=2$.
The extra recurrent state contains $r$ numbers per token and head.

The forward computation uses two attention reads. The first reads
$[n,\operatorname{vec}(nn^\top)]$ to obtain the required statistics,
where $\operatorname{vec}$ flattens a matrix into a vector.
After solving the small covariance system and finding the maximum and
minimum projected scores over the causal prefix, the second read
uses augmented queries and keys
\begin{equation}
 q_i'=[q_i,\sqrt{d_h}\alpha_i],\qquad k_j'=[k_j,n_j]
 \label{eq:augmented}
\end{equation}
with the original $1/\sqrt{d_h}$ scale and concatenated values $[v_j,w_j]$.
Its output is precisely \Cref{eq:jointread}. This form supports exact fused
attention kernels, which avoid storing the full attention matrix
\cite{dao2022flashattention}; processing the sources in blocks computes
the directional range.
For a shared core of $K$ blocks, the number of attention calls per loop
becomes $K+1$. \Cref{app:implementation} details the implementation and
its cost.

We initialize $W_r$ to zero and $W_w$ randomly. The initial attention
distribution is therefore $p$, while the write projection creates targets
that supply a learning signal to $W_r$ on later visits. The model is
optimized with the next-token objective. Training samples a loop depth
per sequence from a Poisson distribution with mean eight, clamped below
at one: $T=\max(1,\operatorname{Poisson}(8))$. It differentiates
through the last $\min(T,4)$ loops. Earlier loops supply their states
without propagating gradients through those earlier computations, a
procedure called truncated backpropagation. Both hidden state and reading
target follow this boundary, and evaluation applies the
shared core for eight recurrent steps.

\begin{algorithm}[t]
 \caption{One MoTA attention visit}
 \label{alg:mota}
 \begin{algorithmic}
  \STATE {\bfseries Input:} query/key/value matrices $Q,K,V$, target $m$
  \STATE $b\gets1$ if a previous target exists, otherwise $b\gets0$
  \STATE Compute read/write coordinates $n,w$ using \Cref{eq:coordinates}
  \STATE $\ell\gets QK^\top/\sqrt{d_h}$; set $\ell_{ij}=-\infty$ for $j>i$
  \STATE $p\gets\softmax(\ell)$
  \STATE $\mu\gets\E_p[n]$, $\Sigma\gets\E_p[nn^\top]-\mu\mu^\top$
  \STATE $d\gets b\,(\Sigma+\lambda I)^{-1}(m-\mu)$
  \STATE $\rho_i\gets\max_{j\leq i}n_j^\top d_i-\min_{j\leq i}n_j^\top d_i$
  \STATE $\alpha\gets\eta(\rho)d$, $a\gets\softmax(\ell+\alpha^\top n)$
  \STATE {\bfseries Return:} $y=\E_a[v]$, $m_{\mathrm{next}}=\E_a[w]$
 \end{algorithmic}
\end{algorithm}

\section{Theoretical Analysis}
\label{sec:theory}

The previous section constructs a read from a target and a set of sources.
We now explain why its geometry is appropriate, how its step changes the
distribution, and how the resulting state can be learned. For one attention
query, let $p_j>0$ be the base probabilities and hold the current source
coordinates and values fixed. All expectations below are over these causal
sources. Proofs appear in \Cref{app:proofs}.

\subsection{A Target Matters Through Source Differences}

The attention ratio between two sources satisfies
\begin{equation}
 \log\frac{a_j}{a_k}=\log\frac{p_j}{p_k}
                       +\alpha^\top(n_j-n_k).
 \label{eq:odds}
\end{equation}
Only differences between source coordinates affect selection. Define
$\mathcal S=\spn\{n_j-n_k:j,k\}$, the space of all linear combinations
of source-coordinate differences. Its orthogonal complement
$\mathcal S^\perp$ contains directions perpendicular to every such
difference. This leads to an exact property of the complete update,
including its step size.

\begin{proposition}[Source-dependent invariance]
 \label{prop:invariance}
 For $u\in\mathcal S^\perp$, let $a(m)$ be the distribution produced by
 \mota\ with target $m$. Then
 \begin{equation}
  a(m+u)=a(m).
  \label{eq:invariance}
 \end{equation}
 The optimum $a^\star$ of \Cref{eq:primal} has the same invariance.
 The attention distribution is also unchanged under a rotation or
 reflection of the coordinates: $n_j\mapsto Rn_j$ and $m\mapsto Rm$
 with $R^\top R=I$.
\end{proposition}

To see the mechanism, $\Sigma u=0$, so the direction changes by
$u/\lambda$. Every source score then changes by the same constant.
Its range $\rho$ remains unchanged, as does the final softmax. Thus, a
request that the sources cannot distinguish does not weaken the update
in directions where they do differ. The orthogonal invariance similarly
shows that selection depends on the source geometry rather than the
orientation assigned to the learned coordinate axes.

This property explains why the step uses the actual score range. The
unit-ball bound $\rho\leq2\norm d$ is valid, but $\norm d$ also includes
components orthogonal to $\mathcal S$. A step based on that norm can
shrink an otherwise useful read in response to an irrelevant target
component. \mota\ measures the change where it affects attention.

\subsection{An Analytic Step for Exponential Reweighting}

Along $f(s)=G(sd)$, the third derivative is the centered third moment
of $z_j=n_j^\top d$. The source range therefore bounds the rate at
which the directional curvature changes:
\begin{equation}
 |f'''(s)|\leq\rho\,\Var_{a_s}(z)\leq\rho f''(s),
 \qquad a_{s,j}\propto p_j e^{s z_j}.
 \label{eq:thirdderivative}
\end{equation}
Here $\Var_a(z)=\E_a[(z-\E_a[z])^2]$ is the variance under $a$.
The inequality bounds the rate of change of curvature by curvature
itself. This property, a directional form of generalized
self-concordance \cite{bach2010selfconcordant}, lets us choose a step
directly from the source scores without a numerical line search.

\begin{theorem}[Descent and probability change]
 \label{thm:descent}
 Let $d$ and $\rho$ be defined by \Cref{eq:direction,eq:range}, and set
 $A=d^\top(\Sigma+\lambda I)d$. Define the scalar function
 \begin{equation}
  \kappa(\rho)=\frac{(1+\rho)\log(1+\rho)-\rho}{\rho^2},
  \qquad\kappa(0)=\tfrac12.
 \end{equation}
The update in \Cref{eq:step} guarantees the following decrease in the dual objective
\begin{equation}
  G(\alpha)-G(0)\leq-A\kappa(\rho),
  \label{eq:descentbound}
 \end{equation}
 Whenever $d\ne0$, the dual objective strictly decreases. The source
 range also controls the change in the odds of selecting one source
 over another:
 \begin{equation}
  \frac{1}{1+\rho}\leq
  \frac{a_j/a_k}{p_j/p_k}\leq1+\rho.
  \label{eq:oddsbound}
 \end{equation}
\end{theorem}

The Newton equations give $f'(0)=-A$ and $f''(0)=A$.
Integrating \Cref{eq:thirdderivative} bounds the change in the objective
along the direction. Minimizing that bound gives
$\eta(\rho)=\log(1+\rho)/\rho$ and \Cref{eq:descentbound}.
At the same time, the range of the added logits becomes exactly
$\log(1+\rho)$, giving \Cref{eq:oddsbound}. Thus, the source score range determines a closed-form step size that decreases the dual objective while bounding changes in pairwise attention odds.

\subsection{From a Target to Content and the Next State}

The previous results describe selection. Its effect on the model follows
by differentiating the weighted outputs. Consider a target near the
base-attention mean, $m=\mu+\delta m$, where $\delta m$ is a small change,
and write $J=(\Sigma+\lambda I)^{-1}$. For two source features $x,z$,
their cross-covariance is
$\Cov_p(x,z)=\E_p[(x-\E_p[x])(z-\E_p[z])^\top]$; we abbreviate
$\Cov_p(x,x)$ as $\Cov_p(x)$. First-order response describes the linear
effect of $\delta m$; the terms $O(\norm{\delta m}^2)$ collect remaining
changes bounded by a constant times its squared norm near zero.

\begin{proposition}[Response of the read and write]
 \label{prop:response}
 The first-order responses of \mota\ are
 \begin{align}
  \delta\E_a[n]&=\Sigma J\,\delta m+O(\norm{\delta m}^2),\label{eq:momentresponse}\\
  \delta y&=\Cov_p(v,n)J\,\delta m+O(\norm{\delta m}^2),\label{eq:contentresponse}\\
  \delta m_{\mathrm{next}}&=\Cov_p(w,n)J\,\delta m
                                  +O(\norm{\delta m}^2).
  \label{eq:stateresponse}
 \end{align}
\end{proposition}

In an eigenvector direction of $\Sigma$ with eigenvalue $\sigma$, the
mean responds with gain $\sigma/(\sigma+\lambda)$: available source
variation supports the requested change, while an absent distinction
has zero response. The content response depends on how the read
coordinates correlate with the original values. The next target depends
on how they correlate with the write coordinates. These two cross
covariances explain the purpose of learning separate projections: the
same selection can retrieve useful content and direct the following
read toward another property. Together, the equations make the entire
read-target-read relation explicit.

The linear response $\Cov_p(v,n)J$ has rank at most $r$, while the read
$y$ retains the original $d_h$-dimensional values. The additional state
therefore parameterizes how selection changes; the source values carry
the content returned to the model. Increasing $r$ expands the space of
controllable responses. A compact state can instead steer a rich content
representation through a small number of learned source distinctions.
This separates the role of state dimension from the capacity of the
underlying representation and motivates the dimension ablation.

\paragraph{Learning from the baseline initialization.}
At $W_r=0$, all read coordinates vanish and $a=p$. A valid incoming
target $m$ gives the output sensitivity to a read-projection perturbation
$\Delta W_r$, where $D_{W_r}y[\Delta W_r]$ denotes the derivative of
$y$ along that perturbation:
\begin{equation}
 D_{W_r}y[\Delta W_r]
   =\frac{1}{2\lambda}\Cov_p(v,\bar v)\Delta W_r m.
 \label{eq:initgradient}
\end{equation}
The attention starts from the base distribution, and the target written
by the random write projection supplies a direction in which the read
projection can learn. This connects the geometry of the update to
end-to-end optimization: the new pathway has a first-order learning
signal as soon as a prior target is present.

\section{Experiments}
\label{sec:experiments}

\begin{table*}[t]
\centering
\caption{\textbf{Pretraining: general language and reasoning capabilities.} Models are trained across FineWeb-Edu budgets of 11.2B, 29.6B, 61.6B, and 104B tokens. Faded rows correspond to the non-looped baseline. For evaluation tasks, CORE-Ext. stands for CORE-Extended, HellaSwag-0 denotes zero-shot accuracy, and ARC-E/C refer to ARC-Easy/Challenge. Parentheses report percentage-point changes relative to the Loop Transformer, where \gain{+} indicates improvement and \loss{-} indicates degradation. Bold text denotes the better score in each paired comparison.}
\label{tab:pretrain}
\small
\setlength{\tabcolsep}{2pt}
\renewcommand{\arraystretch}{1.18}
\resizebox{\linewidth}{!}{%
\begin{tabular}{@{}llcccccccc@{}}
\toprule
\multirow{2}{*}{Size} & \multirow{2}{*}{Model} & \multicolumn{4}{c}{Aggregate and language capabilities} & \multicolumn{4}{c}{Science and commonsense reasoning} \\
\cmidrule(lr){3-6}\cmidrule(lr){7-10}
 &  & \mbox{CORE\,$\uparrow$} & \mbox{CORE-Ext.\,$\uparrow$} & \mbox{LAMBADA\,$\uparrow$} & \mbox{HellaSwag-0\,$\uparrow$} & \mbox{ARC-E\,$\uparrow$} & \mbox{ARC-C\,$\uparrow$} & \mbox{PIQA\,$\uparrow$} & \mbox{BoolQ\,$\uparrow$} \\
\midrule
\rowcolor{referencegray} \cellcolor{white} & \textcolor{referenceink}{Transformer} & \textcolor{referenceink}{13.00} & \textcolor{referenceink}{8.80} & \textcolor{referenceink}{24.24} & \textcolor{referenceink}{32.69} & \textcolor{referenceink}{51.22} & \textcolor{referenceink}{26.54} & \textcolor{referenceink}{60.07} & \textcolor{referenceink}{55.78} \\
\cellcolor{white} & \mbox{Loop Transformer} & 14.04 & 9.67 & 26.94 & 33.95 & 53.03 & 27.90 & 63.28 & \textbf{56.79} \\
\oursrow \cellcolor{white}\multirow{-3}{*}{140M} & \textbf{\mota} & \scorecell{\textbf{16.24}}{\gain{+2.20}} & \scorecell{\textbf{10.74}}{\gain{+1.07}} & \scorecell{\textbf{29.61}}{\gain{+2.68}} & \scorecell{\textbf{36.13}}{\gain{+2.18}} & \scorecell{\textbf{56.06}}{\gain{+3.03}} & \scorecell{\textbf{28.84}}{\gain{+0.94}} & \scorecell{\textbf{64.96}}{\gain{+1.69}} & \scorecell{56.61}{\loss{-0.18}} \\
\midrule
\rowcolor{referencegray} \cellcolor{white} & \textcolor{referenceink}{Transformer} & \textcolor{referenceink}{17.46} & \textcolor{referenceink}{11.71} & \textcolor{referenceink}{34.64} & \textcolor{referenceink}{41.81} & \textcolor{referenceink}{59.05} & \textcolor{referenceink}{32.51} & \textcolor{referenceink}{65.83} & \textcolor{referenceink}{53.33} \\
\cellcolor{white} & \mbox{Loop Transformer} & 20.00 & 12.75 & 35.30 & 43.44 & 62.21 & \textbf{33.28} & 67.41 & 53.88 \\
\oursrow \cellcolor{white}\multirow{-3}{*}{370M} & \textbf{\mota} & \scorecell{\textbf{21.06}}{\gain{+1.06}} & \scorecell{\textbf{13.40}}{\gain{+0.65}} & \scorecell{\textbf{36.46}}{\gain{+1.16}} & \scorecell{\textbf{45.39}}{\gain{+1.95}} & \scorecell{\textbf{63.38}}{\gain{+1.18}} & \scorecell{33.19}{\loss{-0.09}} & \scorecell{\textbf{69.59}}{\gain{+2.18}} & \scorecell{\textbf{56.76}}{\gain{+2.87}} \\
\midrule
\rowcolor{referencegray} \cellcolor{white} & \textcolor{referenceink}{Transformer} & \textcolor{referenceink}{22.42} & \textcolor{referenceink}{14.20} & \textcolor{referenceink}{40.54} & \textcolor{referenceink}{50.10} & \textcolor{referenceink}{65.78} & \textcolor{referenceink}{34.90} & \textcolor{referenceink}{70.02} & \textcolor{referenceink}{49.60} \\
\cellcolor{white} & \mbox{Loop Transformer} & 25.07 & 15.19 & 42.07 & 51.12 & 65.99 & 36.01 & 70.67 & 54.22 \\
\oursrow \cellcolor{white}\multirow{-3}{*}{770M} & \textbf{\mota} & \scorecell{\textbf{27.69}}{\gain{+2.62}} & \scorecell{\textbf{16.93}}{\gain{+1.74}} & \scorecell{\textbf{43.57}}{\gain{+1.49}} & \scorecell{\textbf{54.05}}{\gain{+2.94}} & \scorecell{\textbf{67.47}}{\gain{+1.47}} & \scorecell{\textbf{40.61}}{\gain{+4.61}} & \scorecell{\textbf{72.14}}{\gain{+1.47}} & \scorecell{\textbf{61.62}}{\gain{+7.40}} \\
\midrule
\rowcolor{referencegray} \cellcolor{white} & \textcolor{referenceink}{Transformer} & \textcolor{referenceink}{25.45} & \textcolor{referenceink}{15.90} & \textcolor{referenceink}{41.55} & \textcolor{referenceink}{51.16} & \textcolor{referenceink}{67.63} & \textcolor{referenceink}{39.08} & \textcolor{referenceink}{72.36} & \textcolor{referenceink}{63.36} \\
\cellcolor{white} & \mbox{Loop Transformer} & 28.44 & 17.08 & 43.99 & 56.24 & 68.73 & 40.53 & 72.42 & 64.10 \\
\oursrow \cellcolor{white}\multirow{-3}{*}{1.3B} & \textbf{\mota} & \scorecell{\textbf{32.00}}{\gain{+3.56}} & \scorecell{\textbf{19.09}}{\gain{+2.01}} & \scorecell{\textbf{49.80}}{\gain{+5.80}} & \scorecell{\textbf{60.88}}{\gain{+4.64}} & \scorecell{\textbf{72.60}}{\gain{+3.87}} & \scorecell{\textbf{44.11}}{\gain{+3.58}} & \scorecell{\textbf{74.59}}{\gain{+2.18}} & \scorecell{\textbf{66.48}}{\gain{+2.39}} \\
\bottomrule
\end{tabular}
}
\end{table*}

\begin{table*}[t]
\centering
\caption{\textbf{Lightweight math adaptation: learning to use mathematical knowledge.} FineMath4+ budgets are 0.54B tokens for 140M and 1.07B for the other scales. MMLU-CM, MMLU-AA, and MMLU-HSM denote the College Mathematics, Abstract Algebra, and High School Mathematics subjects of MMLU. BB-Math denotes BIG-Bench Elementary Math QA, Arithmetic tests expression completion, WinoGrande evaluates contextual commonsense reasoning.}
\label{tab:adapt}
\small
\setlength{\tabcolsep}{2pt}
\renewcommand{\arraystretch}{1.18}
\resizebox{\linewidth}{!}{%
\begin{tabular}{@{}llccccccccc@{}}
\toprule
\multirow{2}{*}{Size} & \multirow{2}{*}{Model} & \multicolumn{4}{c}{Mathematical solutions and question answering} & \multicolumn{5}{c}{Mathematical knowledge and contextual reasoning} \\
\cmidrule(lr){3-6}\cmidrule(lr){7-11}
 &  & \mbox{MATH\,$\uparrow$} & \mbox{GSM8K\,$\uparrow$} & \mbox{MathQA (0-shot)\,$\uparrow$} & \mbox{BB-Math\,$\uparrow$} & \mbox{MMLU-CM\,$\uparrow$} & \mbox{MMLU-AA\,$\uparrow$} & \mbox{MMLU-HSM\,$\uparrow$} & \mbox{Arithmetic\,$\uparrow$} & \mbox{WinoGrande\,$\uparrow$} \\
\midrule
\rowcolor{referencegray} \cellcolor{white} & \textcolor{referenceink}{Transformer} & \textcolor{referenceink}{1.68} & \textcolor{referenceink}{0.91} & \textcolor{referenceink}{22.88} & \textcolor{referenceink}{25.18} & \textcolor{referenceink}{17.00} & \textcolor{referenceink}{19.00} & \textcolor{referenceink}{19.26} & \textcolor{referenceink}{0.20} & \textcolor{referenceink}{46.33} \\
\cellcolor{white} & \mbox{Loop Transformer} & \textbf{2.50} & \textbf{1.36} & \textbf{24.25} & 25.25 & \textbf{21.00} & \textbf{22.00} & \textbf{21.85} & \textbf{0.50} & 49.33 \\
\oursrow \cellcolor{white}\multirow{-3}{*}{140M} & \textbf{\mota} & \scorecell{1.84}{\loss{-0.66}} & \scorecell{1.06}{\loss{-0.30}} & \scorecell{23.15}{\loss{-1.11}} & \scorecell{\textbf{27.83}}{\gain{+2.59}} & \scorecell{\textbf{21.00}}{\zerodelta{+0.00}} & \scorecell{\textbf{22.00}}{\zerodelta{+0.00}} & \scorecell{21.11}{\loss{-0.74}} & \scorecell{0.00}{\loss{-0.50}} & \scorecell{\textbf{51.54}}{\gain{+2.21}} \\
\midrule
\rowcolor{referencegray} \cellcolor{white} & \textcolor{referenceink}{Transformer} & \textcolor{referenceink}{2.18} & \textcolor{referenceink}{0.53} & \textcolor{referenceink}{19.97} & \textcolor{referenceink}{24.04} & \textcolor{referenceink}{27.00} & \textcolor{referenceink}{21.00} & \textcolor{referenceink}{22.59} & \textcolor{referenceink}{0.90} & \textcolor{referenceink}{49.01} \\
\cellcolor{white} & \mbox{Loop Transformer} & \textbf{2.66} & 0.76 & 21.14 & 24.79 & 28.00 & \textbf{28.00} & \textbf{27.41} & 0.80 & 50.51 \\
\oursrow \cellcolor{white}\multirow{-3}{*}{370M} & \textbf{\mota} & \scorecell{2.54}{\loss{-0.12}} & \scorecell{\textbf{2.50}}{\gain{+1.74}} & \scorecell{\textbf{24.96}}{\gain{+3.82}} & \scorecell{\textbf{25.79}}{\gain{+1.00}} & \scorecell{\textbf{32.00}}{\gain{+4.00}} & \scorecell{26.00}{\loss{-2.00}} & \scorecell{25.56}{\loss{-1.85}} & \scorecell{\textbf{1.70}}{\gain{+0.90}} & \scorecell{\textbf{51.93}}{\gain{+1.42}} \\
\midrule
\rowcolor{referencegray} \cellcolor{white} & \textcolor{referenceink}{Transformer} & \textcolor{referenceink}{2.58} & \textcolor{referenceink}{2.12} & \textcolor{referenceink}{24.02} & \textcolor{referenceink}{24.04} & \textcolor{referenceink}{13.00} & \textcolor{referenceink}{18.00} & \textcolor{referenceink}{22.59} & \textcolor{referenceink}{1.60} & \textcolor{referenceink}{50.59} \\
\cellcolor{white} & \mbox{Loop Transformer} & 2.94 & 2.43 & 25.43 & \textbf{25.82} & 19.00 & 22.00 & \textbf{25.56} & 2.00 & 54.85 \\
\oursrow \cellcolor{white}\multirow{-3}{*}{770M} & \textbf{\mota} & \scorecell{\textbf{2.96}}{\gain{+0.02}} & \scorecell{\textbf{3.03}}{\gain{+0.61}} & \scorecell{\textbf{26.60}}{\gain{+1.17}} & \scorecell{25.36}{\loss{-0.46}} & \scorecell{\textbf{29.00}}{\gain{+10.00}} & \scorecell{\textbf{32.00}}{\gain{+10.00}} & \scorecell{24.44}{\loss{-1.11}} & \scorecell{\textbf{2.40}}{\gain{+0.40}} & \scorecell{\textbf{56.20}}{\gain{+1.34}} \\
\midrule
\rowcolor{referencegray} \cellcolor{white} & \textcolor{referenceink}{Transformer} & \textcolor{referenceink}{3.34} & \textcolor{referenceink}{1.59} & \textcolor{referenceink}{23.79} & \textcolor{referenceink}{25.31} & \textcolor{referenceink}{23.00} & \textcolor{referenceink}{18.00} & \textcolor{referenceink}{24.07} & \textcolor{referenceink}{2.30} & \textcolor{referenceink}{54.78} \\
\cellcolor{white} & \mbox{Loop Transformer} & 3.50 & 2.12 & 25.09 & 25.51 & 29.00 & 20.00 & 26.30 & 2.80 & 56.91 \\
\oursrow \cellcolor{white}\multirow{-3}{*}{1.3B} & \textbf{\mota} & \scorecell{\textbf{4.36}}{\gain{+0.86}} & \scorecell{\textbf{3.34}}{\gain{+1.21}} & \scorecell{\textbf{27.14}}{\gain{+2.04}} & \scorecell{\textbf{25.82}}{\gain{+0.30}} & \scorecell{\textbf{33.00}}{\gain{+4.00}} & \scorecell{\textbf{33.00}}{\gain{+13.00}} & \scorecell{\textbf{29.63}}{\gain{+3.33}} & \scorecell{\textbf{3.90}}{\gain{+1.10}} & \scorecell{\textbf{61.72}}{\gain{+4.81}} \\
\bottomrule
\end{tabular}
}
\end{table*}

\begin{table*}[t]
\centering
\caption{\textbf{Math midtraining: mathematical and scientific reasoning.} Cumulative FineMath4+ budgets are 3.00B, 3.00B, 5.37B, and 10.85B tokens. MMLU-CM denotes MMLU College Mathematics; BB-Math denotes BIG-Bench Elementary Math QA. ARC-E, ARC-C, PubMedQA, and LSAT-AR cover school science, biomedical questions, and analytical reasoning.}
\label{tab:midtrain}
\small
\setlength{\tabcolsep}{2pt}
\renewcommand{\arraystretch}{1.18}
\resizebox{\linewidth}{!}{%
\begin{tabular}{@{}llccccccccc@{}}
\toprule
\multirow{2}{*}{Size} & \multirow{2}{*}{Model} & \multicolumn{5}{c}{Mathematical solutions and knowledge} & \multicolumn{4}{c}{Scientific and analytical reasoning} \\
\cmidrule(lr){3-7}\cmidrule(lr){8-11}
 &  & \mbox{MATH\,$\uparrow$} & \mbox{GSM8K\,$\uparrow$} & \mbox{MathQA (0-shot)\,$\uparrow$} & \mbox{BB-Math\,$\uparrow$} & \mbox{MMLU-CM\,$\uparrow$} & \mbox{ARC-E\,$\uparrow$} & \mbox{ARC-C\,$\uparrow$} & \mbox{PubMedQA\,$\uparrow$} & \mbox{LSAT-AR\,$\uparrow$} \\
\midrule
\rowcolor{referencegray} \cellcolor{white} & \textcolor{referenceink}{Transformer} & \textcolor{referenceink}{2.84} & \textcolor{referenceink}{1.36} & \textcolor{referenceink}{24.49} & \textcolor{referenceink}{25.24} & \textcolor{referenceink}{19.00} & \textcolor{referenceink}{45.20} & \textcolor{referenceink}{23.98} & \textcolor{referenceink}{48.20} & \textcolor{referenceink}{18.26} \\
\cellcolor{white} & \mbox{Loop Transformer} & \textbf{3.28} & \textbf{2.43} & \textbf{25.96} & \textbf{25.39} & 29.00 & \textbf{48.40} & 24.91 & 50.40 & 22.17 \\
\oursrow \cellcolor{white}\multirow{-3}{*}{140M} & \textbf{\mota} & \scorecell{2.38}{\loss{-0.90}} & \scorecell{1.52}{\loss{-0.91}} & \scorecell{25.13}{\loss{-0.84}} & \scorecell{25.27}{\loss{-0.12}} & \scorecell{\textbf{32.00}}{\gain{+3.00}} & \scorecell{45.71}{\loss{-2.69}} & \scorecell{\textbf{25.26}}{\gain{+0.34}} & \scorecell{\textbf{55.10}}{\gain{+4.70}} & \scorecell{\textbf{26.96}}{\gain{+4.78}} \\
\midrule
\rowcolor{referencegray} \cellcolor{white} & \textcolor{referenceink}{Transformer} & \textcolor{referenceink}{2.28} & \textcolor{referenceink}{0.53} & \textcolor{referenceink}{20.57} & \textcolor{referenceink}{24.00} & \textcolor{referenceink}{21.00} & \textcolor{referenceink}{59.22} & \textcolor{referenceink}{32.42} & \textcolor{referenceink}{50.70} & \textcolor{referenceink}{16.96} \\
\cellcolor{white} & \mbox{Loop Transformer} & 2.72 & 0.76 & 23.12 & 24.54 & 22.00 & \textbf{61.28} & 33.79 & 53.90 & \textbf{20.43} \\
\oursrow \cellcolor{white}\multirow{-3}{*}{370M} & \textbf{\mota} & \scorecell{\textbf{3.32}}{\gain{+0.60}} & \scorecell{\textbf{2.27}}{\gain{+1.52}} & \scorecell{\textbf{26.43}}{\gain{+3.32}} & \scorecell{\textbf{25.64}}{\gain{+1.10}} & \scorecell{\textbf{28.00}}{\gain{+6.00}} & \scorecell{60.19}{\loss{-1.09}} & \scorecell{\textbf{33.87}}{\gain{+0.09}} & \scorecell{\textbf{55.00}}{\gain{+1.10}} & \scorecell{19.57}{\loss{-0.87}} \\
\midrule
\rowcolor{referencegray} \cellcolor{white} & \textcolor{referenceink}{Transformer} & \textcolor{referenceink}{6.16} & \textcolor{referenceink}{2.20} & \textcolor{referenceink}{25.59} & \textcolor{referenceink}{24.33} & \textcolor{referenceink}{18.00} & \textcolor{referenceink}{58.96} & \textcolor{referenceink}{32.94} & \textcolor{referenceink}{38.40} & \textcolor{referenceink}{18.26} \\
\cellcolor{white} & \mbox{Loop Transformer} & 7.08 & \textbf{2.88} & 28.68 & 25.77 & 27.00 & 65.07 & 36.52 & 41.40 & 24.35 \\
\oursrow \cellcolor{white}\multirow{-3}{*}{770M} & \textbf{\mota} & \scorecell{\textbf{7.48}}{\gain{+0.40}} & \scorecell{2.58}{\loss{-0.30}} & \scorecell{\textbf{31.16}}{\gain{+2.48}} & \scorecell{\textbf{25.92}}{\gain{+0.15}} & \scorecell{\textbf{31.00}}{\gain{+4.00}} & \scorecell{\textbf{65.45}}{\gain{+0.38}} & \scorecell{\textbf{38.74}}{\gain{+2.22}} & \scorecell{\textbf{46.10}}{\gain{+4.70}} & \scorecell{\textbf{26.52}}{\gain{+2.17}} \\
\midrule
\rowcolor{referencegray} \cellcolor{white} & \textcolor{referenceink}{Transformer} & \textcolor{referenceink}{10.12} & \textcolor{referenceink}{2.50} & \textcolor{referenceink}{30.25} & \textcolor{referenceink}{26.43} & \textcolor{referenceink}{23.00} & \textcolor{referenceink}{59.51} & \textcolor{referenceink}{31.91} & \textcolor{referenceink}{20.90} & \textcolor{referenceink}{23.91} \\
\cellcolor{white} & \mbox{Loop Transformer} & 13.06 & 3.64 & 31.56 & 29.34 & 25.00 & 65.95 & 36.52 & 33.60 & 27.39 \\
\oursrow \cellcolor{white}\multirow{-3}{*}{1.3B} & \textbf{\mota} & \scorecell{\textbf{14.46}}{\gain{+1.40}} & \scorecell{\textbf{6.75}}{\gain{+3.11}} & \scorecell{\textbf{32.90}}{\gain{+1.34}} & \scorecell{\textbf{32.76}}{\gain{+3.42}} & \scorecell{\textbf{33.00}}{\gain{+8.00}} & \scorecell{\textbf{68.77}}{\gain{+2.82}} & \scorecell{\textbf{39.51}}{\gain{+2.99}} & \scorecell{\textbf{43.70}}{\gain{+10.10}} & \scorecell{\textbf{28.70}}{\gain{+1.30}} \\
\bottomrule
\end{tabular}
}
\end{table*}

\begin{table*}[t]
\centering
\caption{\textbf{State dimension.} $r$ is the number of read/write coordinates and target components per head. 370M models at eight loops. Pretraining uses 29.6B FineWeb-Edu tokens; lightweight adaptation and midtraining use 1.07B and 3B cumulative FineMath4+ tokens.}
\label{tab:dimension}
\small
\setlength{\tabcolsep}{2pt}
\renewcommand{\arraystretch}{1.18}
\resizebox{\linewidth}{!}{%
\begin{tabular}{@{}lcccccccc@{}}
\toprule
 & \multicolumn{2}{c}{Pretraining} & \multicolumn{2}{c}{Lightweight math adaptation} & \multicolumn{4}{c}{Math midtraining} \\
\cmidrule(lr){2-3}\cmidrule(lr){4-5}\cmidrule(l){6-9}
Configuration & \mbox{CORE\,$\uparrow$} & \mbox{CORE-Ext.\,$\uparrow$} & \mbox{MMLU-CM\,$\uparrow$} & \mbox{BB-Math\,$\uparrow$} & \mbox{MATH\,$\uparrow$} & \mbox{GSM8K\,$\uparrow$} & \mbox{MathQA (0-shot)\,$\uparrow$} & \mbox{PubMedQA\,$\uparrow$} \\
\midrule
\rowcolor{referencegray} \textcolor{referenceink}{Loop Transformer} & \textcolor{referenceink}{20.00} & \textcolor{referenceink}{12.75} & \textcolor{referenceink}{28.00} & \textcolor{referenceink}{24.79} & \textcolor{referenceink}{2.72} & \textcolor{referenceink}{0.76} & \textcolor{referenceink}{23.12} & \textcolor{referenceink}{53.90} \\
\oursrow \mota\ ($r=2$) & 21.06 & 13.40 & \textbf{32.00} & \textbf{25.79} & \textbf{3.32} & \textbf{2.27} & \textbf{26.43} & \textbf{55.00} \\
$r=1$ & \scorecell{\textbf{22.11}}{\gain{+1.05}} & \scorecell{13.20}{\loss{-0.21}} & \scorecell{24.00}{\loss{-8.00}} & \scorecell{24.82}{\loss{-0.97}} & \scorecell{2.98}{\loss{-0.34}} & \scorecell{2.12}{\loss{-0.15}} & \scorecell{26.30}{\loss{-0.13}} & \scorecell{49.20}{\loss{-5.80}} \\
$r=4$ & \scorecell{21.53}{\gain{+0.47}} & \scorecell{\textbf{13.68}}{\gain{+0.27}} & \scorecell{21.00}{\loss{-11.00}} & \scorecell{25.47}{\loss{-0.32}} & \scorecell{3.10}{\loss{-0.22}} & \scorecell{1.74}{\loss{-0.53}} & \scorecell{25.93}{\loss{-0.50}} & \scorecell{49.00}{\loss{-6.00}} \\
$r=8$ & \scorecell{20.69}{\loss{-0.37}} & \scorecell{12.89}{\loss{-0.51}} & \scorecell{\textbf{32.00}}{\zerodelta{+0.00}} & \scorecell{24.67}{\loss{-1.12}} & \scorecell{3.30}{\loss{-0.02}} & \scorecell{\textbf{2.27}}{\zerodelta{+0.00}} & \scorecell{25.83}{\loss{-0.60}} & \scorecell{51.70}{\loss{-3.30}} \\
\bottomrule
\end{tabular}
}
\end{table*}

\begin{table*}[t]
\centering
\caption{\textbf{Placement within the shared core.} Only the attention layers of the listed shared-core blocks are replaced. 370M models at 8 loops. Pretraining uses 29.6B FineWeb-Edu tokens, lightweight adaptation and midtraining use 1.07B and 3B FineMath4+ tokens.}
\label{tab:placement}
\small
\setlength{\tabcolsep}{2pt}
\renewcommand{\arraystretch}{1.18}
\resizebox{\linewidth}{!}{%
\begin{tabular}{@{}lcccccccc@{}}
\toprule
 & \multicolumn{2}{c}{Pretraining} & \multicolumn{2}{c}{Lightweight math adaptation} & \multicolumn{4}{c}{Math midtraining} \\
\cmidrule(lr){2-3}\cmidrule(lr){4-5}\cmidrule(l){6-9}
Configuration & \mbox{CORE\,$\uparrow$} & \mbox{CORE-Ext.\,$\uparrow$} & \mbox{MMLU-CM\,$\uparrow$} & \mbox{BB-Math\,$\uparrow$} & \mbox{MATH\,$\uparrow$} & \mbox{GSM8K\,$\uparrow$} & \mbox{MathQA (0-shot)\,$\uparrow$} & \mbox{PubMedQA\,$\uparrow$} \\
\midrule
\rowcolor{referencegray} \textcolor{referenceink}{Loop Transformer} & \textcolor{referenceink}{20.00} & \textcolor{referenceink}{12.75} & \textcolor{referenceink}{28.00} & \textcolor{referenceink}{24.79} & \textcolor{referenceink}{2.72} & \textcolor{referenceink}{0.76} & \textcolor{referenceink}{23.12} & \textcolor{referenceink}{53.90} \\
\oursrow \mota\ ($r=2$) & 21.06 & 13.40 & \textbf{32.00} & 25.79 & \textbf{3.32} & \textbf{2.27} & \textbf{26.43} & \textbf{55.00} \\
First two blocks & \scorecell{20.67}{\loss{-0.39}} & \scorecell{12.55}{\loss{-0.85}} & \scorecell{26.00}{\loss{-6.00}} & \scorecell{25.79}{\zerodelta{+0.00}} & \scorecell{3.14}{\loss{-0.18}} & \scorecell{1.52}{\loss{-0.76}} & \scorecell{25.06}{\loss{-1.37}} & \scorecell{41.40}{\loss{-13.60}} \\
All four blocks & \scorecell{\textbf{22.15}}{\gain{+1.09}} & \scorecell{\textbf{14.51}}{\gain{+1.11}} & \scorecell{23.00}{\loss{-9.00}} & \scorecell{24.93}{\loss{-0.86}} & \scorecell{3.10}{\loss{-0.22}} & \scorecell{1.82}{\loss{-0.45}} & \scorecell{25.19}{\loss{-1.24}} & \scorecell{48.90}{\loss{-6.10}} \\
Last block & \scorecell{20.55}{\loss{-0.51}} & \scorecell{12.90}{\loss{-0.50}} & \scorecell{23.00}{\loss{-9.00}} & \scorecell{\textbf{25.87}}{\gain{+0.08}} & \scorecell{\textbf{3.32}}{\zerodelta{+0.00}} & \scorecell{1.59}{\loss{-0.68}} & \scorecell{25.73}{\loss{-0.70}} & \scorecell{47.00}{\loss{-8.00}} \\
\bottomrule
\end{tabular}
}
\end{table*}

\begin{table*}[t]
\centering
\caption{\textbf{Covariance and target discrepancy.} Diagonal removes covariance off-diagonal entries, and isotropic uses equal mean variance in all directions. No mean subtraction retains full covariance but replaces $m-\mu$ with $m$. 370M models at eight loops. Pretraining uses 29.6B FineWeb-Edu tokens, lightweight adaptation and midtraining use 1.07B and 3B FineMath4+ tokens.}
\label{tab:geometry}
\small
\setlength{\tabcolsep}{2pt}
\renewcommand{\arraystretch}{1.18}
\resizebox{\linewidth}{!}{%
\begin{tabular}{@{}lcccccccc@{}}
\toprule
 & \multicolumn{2}{c}{Pretraining} & \multicolumn{2}{c}{Lightweight math adaptation} & \multicolumn{4}{c}{Math midtraining} \\
\cmidrule(lr){2-3}\cmidrule(lr){4-5}\cmidrule(l){6-9}
Configuration & \mbox{CORE\,$\uparrow$} & \mbox{CORE-Ext.\,$\uparrow$} & \mbox{MMLU-CM\,$\uparrow$} & \mbox{BB-Math\,$\uparrow$} & \mbox{MATH\,$\uparrow$} & \mbox{GSM8K\,$\uparrow$} & \mbox{MathQA (0-shot)\,$\uparrow$} & \mbox{PubMedQA\,$\uparrow$} \\
\midrule
\rowcolor{referencegray} \textcolor{referenceink}{Loop Transformer} & \textcolor{referenceink}{20.00} & \textcolor{referenceink}{12.75} & \textcolor{referenceink}{28.00} & \textcolor{referenceink}{24.79} & \textcolor{referenceink}{2.72} & \textcolor{referenceink}{0.76} & \textcolor{referenceink}{23.12} & \textcolor{referenceink}{53.90} \\
\oursrow \mota\ ($r=2$) & 21.06 & 13.40 & \textbf{32.00} & \textbf{25.79} & \textbf{3.32} & \textbf{2.27} & \textbf{26.43} & \textbf{55.00} \\
Diagonal & \scorecell{21.59}{\gain{+0.52}} & \scorecell{13.15}{\loss{-0.25}} & \scorecell{21.00}{\loss{-11.00}} & \scorecell{24.92}{\loss{-0.87}} & \scorecell{2.90}{\loss{-0.42}} & \scorecell{1.52}{\loss{-0.76}} & \scorecell{25.53}{\loss{-0.90}} & \scorecell{38.60}{\loss{-16.40}} \\
Isotropic & \scorecell{21.69}{\gain{+0.63}} & \scorecell{\textbf{13.79}}{\gain{+0.39}} & \scorecell{28.00}{\loss{-4.00}} & \scorecell{24.88}{\loss{-0.91}} & \scorecell{3.04}{\loss{-0.28}} & \scorecell{1.29}{\loss{-0.99}} & \scorecell{25.29}{\loss{-1.14}} & \scorecell{54.00}{\loss{-1.00}} \\
No mean subtraction & \scorecell{\textbf{21.71}}{\gain{+0.65}} & \scorecell{13.49}{\gain{+0.09}} & \scorecell{24.00}{\loss{-8.00}} & \scorecell{25.33}{\loss{-0.46}} & \scorecell{3.30}{\loss{-0.02}} & \scorecell{1.90}{\loss{-0.38}} & \scorecell{25.63}{\loss{-0.80}} & \scorecell{34.60}{\loss{-20.40}} \\
\bottomrule
\end{tabular}
}
\end{table*}

\begin{table*}[t]
\centering
\caption{\textbf{Target feedback at inference.} At each stage, the same 370M checkpoint is evaluated under all settings. Pretraining uses 29.6B FineWeb-Edu tokens, lightweight adaptation and midtraining use 1.07B and 3B FineMath4+ tokens. Target feedback uses the target written by the previous read to correct attention, and Off disables this correction while retaining hidden-state recurrence.}
\label{tab:feedback}
\small
\setlength{\tabcolsep}{2pt}
\renewcommand{\arraystretch}{1.18}
\resizebox{\linewidth}{!}{%
\begin{tabular}{@{}llcccccccc@{}}
\toprule
 & & \multicolumn{2}{c}{Pretraining} & \multicolumn{2}{c}{Lightweight math adaptation} & \multicolumn{4}{c}{Math midtraining} \\
\cmidrule(lr){3-4}\cmidrule(lr){5-6}\cmidrule(l){7-10}
Loops & Target feedback & \mbox{CORE\,$\uparrow$} & \mbox{CORE-Ext.\,$\uparrow$} & \mbox{MMLU-CM\,$\uparrow$} & \mbox{BB-Math\,$\uparrow$} & \mbox{MATH\,$\uparrow$} & \mbox{GSM8K\,$\uparrow$} & \mbox{MathQA (0-shot)\,$\uparrow$} & \mbox{PubMedQA\,$\uparrow$} \\
\midrule
\oursrow 8 & On & 21.06 & \textbf{13.40} & \textbf{32.00} & 25.79 & \textbf{3.32} & \textbf{2.27} & \textbf{26.43} & 55.00 \\
8 & Off & \scorecell{\textbf{21.14}}{\gain{+0.07}} & \scorecell{13.22}{\loss{-0.18}} & \scorecell{26.00}{\loss{-6.00}} & \scorecell{\textbf{25.90}}{\gain{+0.11}} & \scorecell{3.26}{\loss{-0.06}} & \scorecell{1.97}{\loss{-0.30}} & \scorecell{26.20}{\loss{-0.23}} & \scorecell{\textbf{55.20}}{\gain{+0.20}} \\
\bottomrule
\end{tabular}
}
\end{table*}

\begin{table*}[t]
\centering
\caption{\textbf{Loop depth at inference.} At each stage, the same 370M checkpoint is evaluated under all settings. Pretraining uses 29.6B FineWeb-Edu tokens, lightweight adaptation and midtraining use 1.07B and 3B cumulative FineMath4+ tokens. Target feedback uses the target written by the previous read to correct attention, Off disables this correction while retaining hidden-state recurrence.}
\label{tab:depth}
\small
\setlength{\tabcolsep}{2pt}
\renewcommand{\arraystretch}{1.18}
\resizebox{\linewidth}{!}{%
\begin{tabular}{@{}llcccccccc@{}}
\toprule
 & & \multicolumn{2}{c}{Pretraining} & \multicolumn{2}{c}{Lightweight math adaptation} & \multicolumn{4}{c}{Math midtraining} \\
\cmidrule(lr){3-4}\cmidrule(lr){5-6}\cmidrule(l){7-10}
Loops & Target feedback & \mbox{CORE\,$\uparrow$} & \mbox{CORE-Ext.\,$\uparrow$} & \mbox{MMLU-CM\,$\uparrow$} & \mbox{BB-Math\,$\uparrow$} & \mbox{MATH\,$\uparrow$} & \mbox{GSM8K\,$\uparrow$} & \mbox{MathQA (0-shot)\,$\uparrow$} & \mbox{PubMedQA\,$\uparrow$} \\
\midrule
2 & On & \scorecell{16.92}{\loss{-4.15}} & \scorecell{11.53}{\loss{-1.87}} & \scorecell{23.00}{\loss{-9.00}} & \scorecell{25.55}{\loss{-0.24}} & \scorecell{2.64}{\loss{-0.68}} & \scorecell{2.05}{\loss{-0.23}} & \scorecell{25.66}{\loss{-0.77}} & \scorecell{51.80}{\loss{-3.20}} \\
4 & On & \scorecell{20.36}{\loss{-0.70}} & \scorecell{13.31}{\loss{-0.09}} & \scorecell{30.00}{\loss{-2.00}} & \scorecell{\textbf{25.87}}{\gain{+0.08}} & \scorecell{2.98}{\loss{-0.34}} & \scorecell{2.20}{\loss{-0.08}} & \scorecell{26.03}{\loss{-0.40}} & \scorecell{\textbf{55.20}}{\gain{+0.20}} \\
\oursrow 8 & On & 21.06 & \textbf{13.40} & \textbf{32.00} & 25.79 & \textbf{3.32} & \textbf{2.27} & 26.43 & 55.00 \\
16 & On & \scorecell{\textbf{21.15}}{\gain{+0.09}} & \scorecell{13.17}{\loss{-0.23}} & \scorecell{30.00}{\loss{-2.00}} & \scorecell{25.86}{\gain{+0.07}} & \scorecell{3.24}{\loss{-0.08}} & \scorecell{2.05}{\loss{-0.23}} & \scorecell{\textbf{26.70}}{\gain{+0.27}} & \scorecell{55.10}{\gain{+0.10}} \\
\bottomrule
\end{tabular}
}
\end{table*}

Our experiments follow the progression from learning general representations
to using them for mathematical and scientific tasks. We first evaluate
pretraining quality, then examine lightweight math adaptation and math
midtraining. Architectural ablations test the source geometry and placement
that define the attention operator. Inference interventions then examine
how the trained model uses feedback and additional recurrent computation
by changing its forward pass without retraining its parameters.

\subsection{Experimental Setup}

\paragraph{Models and training.}
We use the 140M, 370M, 770M, and 1.3B Parcae families, where M and B denote millions and billions of parameters,
\cite{prairie2026parcae}. The baseline is the corresponding looped
Transformer. \mota\ changes the first attention layer of its shared core
as described in \Cref{sec:implementation}. We also include a fixed-depth Transformer baseline at each model scale to compare standard and recurrent architectures. Following the pretraining setup of Parcae, we use the FineWeb-Edu dataset \cite{penedo2024fineweb}, with token budgets of 11.2B, 29.6B, 61.6B, and 104B across the four model scales, respectively.
All models use a 2,048-token context and next-token prediction.
Mathematical training uses FineMath4+, the FineMath subset with a quality score of at least four \cite{benallal2025smollm2}.
The lightweight adaptation budgets are 0.54B tokens for 140M and 1.07B
for the other scales. Math midtraining is evaluated at cumulative budgets
of 3.00B, 3.00B, 5.37B, and 10.85B tokens. Each baseline-\mota\ pair
uses the same mathematical budget at a reported stage.
Exact configurations, checkpoint sources, and token counts appear in
\Cref{app:implementation,app:evaluation}.

\paragraph{Capability measures.}
General evaluation includes CORE and CORE-Extended, Parcae's aggregate
task scores normalized relative to chance accuracy, alongside language
completion, commonsense reasoning, and science question answering (QA):
LAMBADA, HellaSwag, PIQA, BoolQ, and the ARC tasks
\cite{paperno2016lambada,zellers2019hellaswag,bisk2020piqa,clark2019boolq,clark2018arc}.
PIQA tests physical commonsense, BoolQ tests yes/no question answering,
and ARC is the AI2 Reasoning Challenge, with Easy and Challenge splits.
For mathematics, we use MATH competition problems and GSM8K grade-school
word problems for solution generation,
MathQA, BIG-Bench Elementary Math QA, arithmetic, and the College
Mathematics, Abstract Algebra, and High School Mathematics subjects of
Massive Multitask Language Understanding (MMLU)
\cite{hendrycks2021math,cobbe2021gsm8k,amini2019mathqa,srivastava2023bigbench,hendrycks2021mmlu}.
WinoGrande tests contextual commonsense reasoning after lightweight
adaptation \cite{sakaguchi2020winogrande}.
We evaluate the midtrained models on ARC for school science and PubMedQA
for biomedical knowledge \cite{clark2018arc,jin2019pubmedqa}.
The Law School Admission Test analytical-reasoning task (LSAT-AR),
evaluated through AGIEval, extends this panel to analytical reasoning
\cite{zhong2023agieval}. The structural ablations use the same stage-specific
benchmarks as the main tables, connecting each component to the reported
general, mathematical, and scientific capabilities.
These tasks test different consequences of improved internal reading:
recovering learned knowledge, combining relations, and generating a solution.
Zero-shot evaluation supplies no solved examples in the prompt; $k$-shot
evaluation supplies $k$ demonstrations. Individual-task accuracies and
normalized aggregates are multiplied by 100, with larger values better.

\subsection{Pretraining: Improving the General Representation}

\Cref{tab:pretrain} shows that the recurrent attention modification
improves general model quality before specialized mathematical training.
CORE increases at every scale, with gains of 2.20, 1.06, 2.62, and 3.56
points. CORE-Extended follows the same pattern. The improvement extends
across language completion and contextual reasoning: LAMBADA, HellaSwag,
ARC-Easy, and PIQA all improve at each of the four scales.
At 770M and 1.3B, \mota\ also improves ARC-Challenge and BoolQ.

These results establish the starting point for the subsequent stages.
The architecture learns a stronger general representation from web text,
so the benefit of its reading mechanism already appears under the
ordinary language-modeling objective. The larger models show the
clearest combination of gains in language completion, commonsense,
and scientific knowledge.

\subsection{Lightweight Math Adaptation: Using Learned Knowledge}

The adaptation results in \Cref{tab:adapt} examine how quickly the
pretrained models become useful on mathematical tasks. At 770M,
\mota\ improves College Mathematics and Abstract Algebra by ten points
each. At 1.3B, the corresponding gains are four and thirteen points,
alongside improvements in GSM8K, MathQA, BB-Math, and arithmetic.
At 370M, the improvements include GSM8K, MathQA, College Mathematics,
and arithmetic. The 140M results are more mixed, with the clearest gain
in BB-Math. The expanded adaptation panel also includes solution
generation on MATH and MMLU High School Mathematics.
At 1.3B, the gains on MATH and High School Mathematics are 0.86 and
3.33 points, respectively. WinoGrande improves at all four scales,
by 2.21, 1.42, 1.34, and 4.81 points, extending the adaptation-stage
gains to reasoning about contextual relations.

The larger-scale pattern connects naturally to the architectural
motivation. A compact target changes how the recurrent model selects
from its current representations, while the read and write projections
are learned from the same prediction signal as the backbone. The
adaptation stage reveals the value of that architecture when training
begins to emphasize mathematical relations. In particular, the gains
in algebra and college mathematics show that the improvement is visible
in knowledge-intensive questions as well as numerical tasks.

\subsection{Math Midtraining: Solutions and Scientific Reasoning}

\Cref{tab:midtrain} evaluates the models after a larger mathematical
training budget. College Mathematics and PubMedQA improve at every
scale. At 770M, \mota\ improves MathQA by 2.48 points, PubMedQA by
4.70 points, and LSAT-AR by 2.17 points. At 1.3B, it improves
MATH by 1.40 points and GSM8K by 3.11 points, together with gains in
MathQA, BB-Math, College Mathematics, PubMedQA, and LSAT-AR.

The expanded science panel reveals where this transfer is strongest.
At 1.3B, \mota\ gains 2.82 points on ARC-Easy, 2.99 on ARC-Challenge,
and 10.10 on PubMedQA. ARC-Challenge improves at every
scale. These gains extend across school science and biomedical
questions, with the strongest combined improvement at the largest scale.

The three stages expose complementary aspects of the same architecture.
General pretraining strengthens the representation; lightweight
adaptation makes mathematical knowledge more usable; and further
midtraining improves solution generation, scientific question answering,
and analytical reasoning. This progression gives the learned
reading target a concrete role in the empirical story: it changes
an internal selection process whose benefits appear under general
training and persist into specialized reasoning.

\subsection{Which Parts of the Operator Matter?}

\paragraph{General gains establish the basis for the design tradeoff.}
The faded baseline rows in \Cref{tab:dimension,tab:placement,tab:geometry}
carry over the 370M references from the main tables. Full \mota\ raises
pretraining CORE from 20.00 to 21.06 and CORE-Extended from 12.75 to
13.40. Its gains continue through mathematical training: at midtraining,
MathQA reaches 26.43 versus the baseline's 23.12, and GSM8K reaches
2.27 versus 0.76. The tradeoff arises among \mota\ configurations on top of
these baseline improvements. For example, replacing all four core
attentions raises pretraining CORE to 22.15 and CORE-Extended to 14.51;
the first-block design instead achieves higher midtraining MathQA
(26.43 versus 25.19) and PubMedQA (55.00 versus 48.90).
We favor the configuration that combines improved general pretraining
with stronger mathematical learning and scientific question answering.

\paragraph{Covariance describes how the sources can change.}
The geometry variants in \Cref{tab:geometry} replace the full covariance
with a diagonal approximation, which removes correlations, or an
isotropic approximation, which assigns the same average variance to
every coordinate. A third variant, \emph{No mean subtraction}, retains
the full covariance but uses $m$ in place of $m-\mu$ in the correction.
The complete operator gives the strongest
MATH, GSM8K, MathQA, and PubMedQA results within this group. On GSM8K, the full,
diagonal, and isotropic variants score 2.27, 1.52, and 1.29,
respectively; on MATH they score 3.32, 2.90, and 3.04.
Full covariance also gives 55.00 on PubMedQA, compared with 38.60
for the diagonal approximation and 34.60 without mean subtraction.
This comparison connects to \Cref{eq:momentresponse}: correlations
between source properties participate in translating the target into
a useful change of attention. Subtracting $\mu$ makes a distinct
contribution: it compares the requested read with what base attention
already returns. The complete direction is zero when $m=\mu$, whereas
$(\Sigma+\lambda I)^{-1}m$ can still request a change. Together, the
covariance and the target discrepancy determine how the available
sources should be reweighted to meet the next reading goal.

\paragraph{State capacity and placement serve different purposes.}
\Cref{tab:dimension,tab:placement} compare state dimensions and core layers.
Increasing the state dimension does not uniformly improve mathematical
performance. At the midtraining endpoint, the two-dimensional state
gives the best MATH and MathQA results among the dimension variants.
During lightweight adaptation, it ties the highest College Mathematics
score and leads BB-Math within this group. Placement produces a related distinction: replacing
all four core attentions gives the strongest pretraining CORE and
CORE-Extended scores, whereas modifying only the first gives stronger
midtraining GSM8K, MathQA, and PubMedQA. The compact state controls the
source mixture while the original values carry the full content, as
\Cref{eq:contentresponse} makes explicit. Placing this control early
in each loop lets the remaining blocks process the selected information
before the next visit.

\paragraph{Mathematical training reveals the value of the learned geometry.}
At 1.07B tokens, full \mota\ reaches 32.00 on College Mathematics,
compared with 21.00 for diagonal covariance, 28.00 for the isotropic
approximation, and 24.00 without mean subtraction. Full \mota\ also leads
BB-Math against each of these three variants.
The same-task comparison across stages is particularly informative:
full and diagonal covariance score 17.00 and 31.00 on College Mathematics
after pretraining, but 32.00 and 21.00 after lightweight adaptation.
Their ranking reverses as training begins to emphasize mathematical
relations. This connects the stage comparison to \Cref{eq:contentresponse},
where the effect of a target on retrieved content is determined by
learned read-value correlations. The source coordinates and write targets
are trained jointly with the backbone, giving mathematical training a
direct way to shape the model's subsequent selections.

\subsection{How Does the Model Use Recurrent Computation?}

\paragraph{Feedback helps the model use its mathematical knowledge.}
In \Cref{tab:feedback}, we keep the trained 370M parameters and eight-loop computation fixed while varying whether a previous read supplies a target for the next selection.
Both settings retain the backbone learned through
\mota\ training and the recurrent hidden-state path. Enabling feedback
raises adapted College Mathematics from 26.00 to 32.00, above the
baseline's 28.00, and improves midtraining MATH, GSM8K, and MathQA by
0.06, 0.30, and 0.23 points. The setting without feedback is ahead by
0.07 on pretraining CORE, 0.11 on adapted BB-Math, and 0.20 on
midtraining PubMedQA. The clearest contribution is thus to adapted
College Mathematics, alongside gains on the three midtraining
mathematical tasks. In \Cref{eq:contentresponse}, learned read-value
correlations govern the local change in retrieved content as the target
changes. This connects the intervention to the mechanism: feedback
directs which information the model selects for a query, while the
trained backbone continues to carry its learned capabilities.
The nine case studies in Appendix~\ref{app:attentioncases} complement this
analysis by visualizing attention across recurrent loops at matched
reasoning prefixes. They make the next-read motivation concrete:
different intermediate computations call for different source conditions.

\paragraph{Loop depth changes the balance of capabilities.}
\Cref{tab:depth} evaluates the same checkpoint within each stage at
different inference depths. Increasing depth from two to eight loops
raises pretraining CORE from 16.92 to 21.06, adapted College Mathematics
from 23.00 to 32.00, and midtraining MATH from 2.64 to 3.32.
At sixteen loops, CORE and MathQA rise further to 21.15 and 26.70,
while eight loops lead on CORE-Extended, adapted College Mathematics,
MATH, and GSM8K. Each loop updates both the hidden state and the next
reading target, extending the learned sequence of selections.
The resulting trade-off makes inference depth a choice of computation
and task performance: eight loops provide a strong default at lower
cost, and sixteen offer further gains on selected tasks.

\paragraph{Architectural cost.}
With $r=2$, \mota\ adds 3,072 to 6,144 parameters across the four
families, and two state scalars per token and head. It adds one attention
read to compute moments and a maximum/minimum score calculation to the
first core block.
For cores with two, four, six, and eight blocks, the attention-call
ratios are $3/2$, $5/4$, $7/6$, and $9/8$, respectively.
The modification therefore concentrates its extra computation in one
shared operator while preserving the surrounding architecture.

\FloatBarrier
\section{Conclusion}
\label{sec:conclusion}

We introduced \mota, an attention architecture that lets one read specify
a target for the next loop. A soft moment objective connects that target
to the current sources, yielding a covariance-based direction and an
analytic step determined by source-score differences. The same attention
distribution returns content and writes the next target, making successive
reads part of one learned recurrent computation. The analysis explains
how this computation respects source geometry and learns from a baseline
initialization. Across pretraining, lightweight math adaptation, and math
midtraining, the experiments show improved general model quality and
stronger mathematical and scientific capabilities. These results identify
the connection between successive attention reads as a productive
architectural direction for looped language models.

\section*{Impact Statement}
This work studies how language-model architectures can better learn and use
mathematical and scientific information. Improved reasoning can support
education and scientific work, while incorrect generated solutions can also
mislead readers. The architecture is trained on general and mathematical text
and evaluated on established benchmarks. Its applications inherit the broader
benefits and risks of generative language models.

\setlength{\parfillskip}{0pt plus 1fil}
\setlength{\bibsep}{0.5pt plus 0.5pt minus 0.25pt}
\bibliography{references}
\bibliographystyle{icml2026}
\clearpage
\appendix
\onecolumn
\setcounter{table}{0}
\setcounter{figure}{0}
\renewcommand{\thetable}{S\arabic{table}}
\renewcommand{\thefigure}{S\arabic{figure}}
\renewcommand{\theHtable}{supp.\arabic{table}}
\renewcommand{\theHfigure}{supp.\arabic{figure}}
\setlength{\parfillskip}{0pt plus .35\linewidth}
\section{Derivations and Proofs}
\label{app:proofs}

\subsection{Soft Moment Matching and Its Dual}
\label{app:duality}

Let $s$ be the number of causal sources,
$N=[n_1,\ldots,n_s]\in\R^{r\times s}$, and let $p$ have positive entries.
The objective
$F(a)=\lambda\KL(a\Vert p)+\tfrac12\norm{Na-m}^2$ is strictly convex on
the simplex, so its minimizer is unique. The entropy term makes every
weight at this minimizer positive. Expressing the quadratic as a maximum
over an auxiliary vector $\alpha$ gives
\begin{align}
 \min_{a\in\Delta}\frac{F(a)}{\lambda}
 &=\min_{a\in\Delta}\max_{\alpha\in\R^r}
   \left\{\KL(a\Vert p)+\alpha^\top(m-Na)
                     -\frac{\lambda}{2}\norm{\alpha}^2\right\}\\
 &=\max_{\alpha\in\R^r}
   \left\{m^\top\alpha-\frac{\lambda}{2}\norm{\alpha}^2
                     -\log\sum_jp_je^{\alpha^\top n_j}\right\}
 =-\min_\alpha G(\alpha).
 \label{eq:dualproof}
\end{align}
The inner entropy minimization gives
$a_j(\alpha)=p_je^{\alpha^\top n_j}/\sum_kp_ke^{\alpha^\top n_k}$.
The mean $\mu_\alpha=\E_{a(\alpha)}[n]$ and covariance
$\Sigma_\alpha=\Cov_{a(\alpha)}(n)$ determine the derivatives of the dual:
\begin{equation}
 \nabla G(\alpha)=\mu_\alpha+\lambda\alpha-m,
 \qquad \nabla^2G(\alpha)=\Sigma_\alpha+\lambda I.
 \label{eq:dualderivatives}
\end{equation}
The log-partition function is the log normalizer
$\log\sum_jp_je^{\alpha^\top n_j}$; its first and second derivatives
are the mean and covariance above \cite{wainwright2008graphical}.
The same covariance is the \emph{Fisher information}, which measures
how the distribution responds to changes in $\alpha$: the log-probability
gradient is $g_j=\nabla_\alpha\log a_j=n_j-\mu_\alpha$, and
$\E_{a(\alpha)}[g_jg_j^\top]=\Sigma_\alpha$.
Adding the ridge bounds the Hessian below by $\lambda I$, making $G$
strongly convex with a unique minimizer. At the optimum,
$m-\E_{a^\star}[n]=\lambda\alpha^\star$: the residual target discrepancy
is the regularized score correction. The forward operator evaluates one
analytic step from $\alpha=0$, where the moments are available from base
attention.

\subsection[Proof of source-dependent invariance]{Proof of \Cref{prop:invariance}}

\begin{proof}
For $u\in\mathcal S^\perp$, there is a scalar $c$ such that
$n_j^\top u=c$ for every $j$. Therefore
$(n_j-\mu)^\top u=0$ and $\Sigma u=0$. Under $m'=m+u$,
\begin{equation}
 d'=(\Sigma+\lambda I)^{-1}(m+u-\mu)=d+u/\lambda,
 \qquad n_j^\top d'=n_j^\top d+c/\lambda.
\end{equation}
The projected score range is unchanged, so $\eta(\rho')=\eta(\rho)$.
The final logits differ by the query-constant shift
$\eta(\rho)c/\lambda$, which cancels in softmax.
For the primal problem,
\begin{equation}
 \norm{Na-m-u}^2-\norm{Na-m}^2
 =-2c+2u^\top m+\norm u^2,
\end{equation}
which is independent of $a$; its minimizing distribution is unchanged.
Finally, under an orthogonal $R$, we have $\mu'=R\mu$,
$\Sigma'=R\Sigma R^\top$, and $d'=Rd$.
Thus $n_j'^\top d'=n_j^\top d$, giving identical range and logits.
If write coordinates are rotated as well, $m_{\mathrm{next}}'=Rm_{\mathrm{next}}$,
so the state representation transforms consistently through the loop.
\end{proof}

\subsection[Proof of descent and probability change]{Proof of \Cref{thm:descent}}

\begin{proof}
Let $f(s)=G(sd)$, $z_j=n_j^\top d$, and
$A=d^\top(\Sigma+\lambda I)d$. The Newton equation implies
\begin{equation}
 f'(0)=-A,\qquad f''(0)=A,\qquad
 f''(s)=\Var_{a_s}(z)+\lambda\norm d^2,
\end{equation}
and the third derivative is
$f'''(s)=\E_{a_s}[(z-\E_{a_s}z)^3]$.
The mean of $z$ lies between its smallest and largest source values,
an interval of width $\rho$.
It follows that
\begin{equation}
 |f'''(s)|\leq\E_{a_s}|z-\E_{a_s}z|^3
              \leq\rho\Var_{a_s}(z)\leq\rho f''(s).
\end{equation}
For $s\geq0$, integration yields
\begin{align}
 f''(s)&\leq Ae^{\rho s},\\
 f'(s)&\leq-A+A\frac{e^{\rho s}-1}{\rho},\\
 f(s)-f(0)&\leq-As+
       A\frac{e^{\rho s}-1-\rho s}{\rho^2}.
 \label{eq:lineupper}
\end{align}
The upper bound is minimized at
$s=\log(1+\rho)/\rho$. Substitution gives
$f(s)-f(0)\leq-A[(1+\rho)\log(1+\rho)-\rho]/\rho^2$.
The numerator is positive for $\rho>0$, since its derivative is
$\log(1+\rho)$ and its value at zero is zero. For $\rho=0$, the
log-partition is linear along $d$ and $f$ is quadratic;
$s=1$ gives the exact change $-A/2$.
Lastly,
\begin{equation}
 \max_{j,k}\alpha^\top(n_j-n_k)=\eta(\rho)\rho=\log(1+\rho).
\end{equation}
Exponentiating the pairwise log-odds identity gives
\Cref{eq:oddsbound}.
\end{proof}

\paragraph{Comparison with a unit-ball step.}
Since $\norm{n_j}\leq1$, the bound $\rho\leq2\norm d$ holds.
The function $\eta$ is decreasing on $[0,\infty)$, so
$\eta(2\norm d)\leq\eta(\rho)$.
The derivative bound above is nonpositive throughout
$0\leq s\leq\eta(\rho)$. Consequently,
\begin{equation}
 G\big(\eta(\rho)d\big)\leq G\big(\eta(2\norm d)d\big).
\end{equation}
The source-dependent step can therefore use the available direction more
fully while retaining the same descent argument.

\subsection[Read and write response; initialization sensitivity]{Proof of \Cref{prop:response} and Initialization Sensitivity}

\begin{proof}
For $m=\mu+\delta m$, the direction is $d=J\delta m$.
Because $\rho=O(\norm{\delta m})$ and
$\eta(\rho)=1-\rho/2+O(\rho^2)$, we have
$\alpha=J\delta m+O(\norm{\delta m}^2)$.
Differentiating the exponential family at zero gives
\begin{equation}
 a_j=p_j+p_j(n_j-\mu)^\top J\delta m
                       +O(\norm{\delta m}^2).
\end{equation}
Multiplication by $n_j$, $v_j$, or $w_j$ and summation over $j$ gives
\Cref{eq:momentresponse,eq:contentresponse,eq:stateresponse}.
\end{proof}

To obtain \Cref{eq:initgradient}, vary the read projection as
$W_r=\epsilon\Delta W_r$, holding the other inputs of the attention
operator fixed. The Jacobian of $\phi$ at zero is the identity,
$D\phi(0)=I$, so
\begin{equation}
 n_j=\frac{\epsilon}{2}\Delta W_r^\top\bar v_j+O(\epsilon^3),
 \quad\mu=O(\epsilon),\quad\Sigma=O(\epsilon^2),
 \quad d=m/\lambda+O(\epsilon).
\end{equation}
The score range is $O(|\epsilon|)$, giving
\begin{equation}
 \alpha^\top n_j
 =\frac{\epsilon}{2\lambda}\bar v_j^\top\Delta W_r m
   +O(\epsilon^2).
\end{equation}
Differentiating softmax and summing the content values yields
$D_{W_r}y[\Delta W_r]=(2\lambda)^{-1}
\Cov_p(v,\bar v)\Delta W_r m$.

\subsection{A Two-Source Example}
\label{app:example}

Consider equal base weights, $n_1=(-0.4,0.5)$,
$n_2=(0.4,0.5)$, $\lambda=0.1$, and target $m=(0.2,m_2)$.
Then $\mu=(0,0.5)$, $\Sigma=\operatorname{diag}(0.16,0)$, and
\begin{equation}
 d=\left(\frac{0.2}{0.26},\frac{m_2-0.5}{0.1}\right),
 \qquad \rho=0.8\frac{0.2}{0.26}.
\end{equation}
The second coordinate is identical across sources. Varying its target
therefore changes neither the desired source mixture nor the \mota\ read.
The following probabilities are calculated from these equations:
\begin{center}
\begin{tabular}{lccc}
\toprule
Target coordinate & Unit-ball step & \mota & Exact moment optimum\\
\midrule
$m_2=+0.5$ & 0.592093 & 0.617647 & 0.651941\\
$m_2=-0.5$ & 0.523355 & 0.617647 & 0.651941\\
\bottomrule
\end{tabular}
\end{center}
Each entry is the probability assigned to $n_2$. The exact moment
optimum is the minimizer of \Cref{eq:primal}. The unit-ball step
uses $\eta(2\norm d)d$. Its response weakens when the target changes in
a direction that the source distribution cannot alter, illustrating
the distinction captured by \Cref{prop:invariance}.

\section{Model Configuration and Implementation}
\label{app:implementation}

\begin{table}[H]
\centering
\caption{\textbf{Model configurations.} Prelude/core/coda entries give block counts. All attention heads have width 128. Family labels follow Parcae; parameter counts include the added projections.}
\label{tab:modelconfig}
\small
\setlength{\tabcolsep}{3pt}
\renewcommand{\arraystretch}{1.18}
\begin{tabularx}{\linewidth}{@{}llCCCC@{}}
\toprule
Family & Width & Heads & Prelude/core/coda & \mota\ parameters & Added parameters \\
\midrule
140M & 768 & 6 & 2/2/2 & 144,326,208 & 3,072 \\
370M & 1024 & 8 & 4/4/4 & 388,007,424 & 4,096 \\
770M & 1280 & 10 & 6/6/6 & 776,660,800 & 5,120 \\
1.3B & 1536 & 12 & 8/8/8 & 1,338,597,888 & 6,144 \\
\bottomrule
\end{tabularx}
\end{table}

The source range is evaluated over the complete causal prefix. Processing
sources in blocks and retaining their running minimum and maximum avoids
storing an additional dense attention matrix. The first attention read
uses $r+r^2$ moment features;
at $r=2$ these are the two means and four entries of the second moment.
The statistics are accumulated by fused attention. In 32-bit floating
point (FP32), the recovered covariance is averaged with its transpose
to ensure symmetry, then used in a closed-form $2\times2$ linear solve.
The second read combines $[v,w]$ and uses the original head scale
$1/\sqrt{d_h}$. The implementation subtracts the first source coordinate
from each augmented key, $n_j-n_1$, which adds a common logit shift and
leaves the distribution unchanged. For $d_h=128$, the concatenated
vectors are zero-padded to 136 channels to meet the kernel's layout.

For a sequence of length $L$, the source-range computation requires
$O(HL^2r)$ arithmetic and $O(HL)$ retained reduction information.
The learned projections add $2d_{\mathrm{model}}r$ parameters; the recurrent target
contains $HLr$ scalars per sequence. The extra attention read is confined
to the modified block. With $K=2,4,6,8$ shared blocks in the four model
families, the respective attention-call ratios are
$3/2$, $5/4$, $7/6$, and $9/8$ per loop. Actual runtime also depends on
finding score extrema, the extra moment features, and the execution backend.

Read and write projections use AdamW without weight decay, while the
backbone follows the Parcae MuonAdamW recipe: Muon updates selected
matrix weights, while AdamW updates the remaining parameters.
Targets are reset at the start of a model forward and carried across
its loops; read and write coordinates are recomputed from each loop's values.
The validity mask distinguishes a missing target on the first visit from
a valid target at zero. At the truncation boundary, both target and
hidden state are detached: their values are retained, but gradients
do not propagate into the earlier loops.

\section{Evaluation Setup and Inference Ablations}
\label{app:evaluation}

\begin{table}[H]
\centering
\caption{\textbf{Training and evaluation budgets.} FineWeb-Edu follows the Parcae-aligned recipe. Both mathematical budgets are cumulative FineMath4+ tokens, measured in billions. Updates count from the start of mathematical training.}
\label{tab:budgets}
\small
\setlength{\tabcolsep}{3pt}
\renewcommand{\arraystretch}{1.18}
\begin{tabularx}{\linewidth}{@{}llCCCC@{}}
\toprule
Family & \headcell{Pretraining\\tokens (B)} & \headcell{Adaptation\\tokens (B)} & \headcell{Midtraining\\tokens (B)} & \headcell{Adaptation\\updates} & \headcell{Midtraining\\updates} \\
\midrule
140M & 11.2 & 0.536871 & 2.999976 & 1024 & 5722 \\
370M & 29.6 & 1.073742 & 2.999976 & 2048 & 5722 \\
770M & 61.6 & 1.073742 & 5.368709 & 2048 & 10240 \\
1.3B & 104 & 1.073742 & 10.849092 & 2048 & 20693 \\
\bottomrule
\end{tabularx}
\end{table}

The three evaluation stages use FineWeb-Edu pretraining followed by
FineMath4+ training. Lightweight math adaptation and math midtraining
refer to the early and later cumulative mathematical budgets in
\Cref{tab:budgets}. The later budget includes the earlier one; the two
mathematical evaluations follow the same continuation schedule.

Pretraining baseline results use the author-released Parcae models.
The 140M and 1.3B mathematical baselines continue the released weights;
the 370M and 770M mathematical baselines use our pretrained backbone
controls. These baselines are labeled Loop Transformer in the tables.
CORE and CORE-Extended pretraining values for both Transformer baselines
are taken from Parcae's published results. The fixed-depth Transformer
individual-task accuracies come from companion experiments and are
computed from correct-answer counts. Individual tasks for Loop Transformer
and all \mota\ results are evaluated in our evaluation pipeline. Local CORE evaluation uses
text-only prompt encoding and the upstream task definitions and aggregation.
For a task set $\mathcal T$, this aggregate is
\begin{equation}
 \frac{100}{|\mathcal T|}\sum_{\tau\in\mathcal T}
 \frac{a_\tau-b_\tau}{1-b_\tau},
\end{equation}
where $a_\tau$ is task accuracy and $b_\tau$ is its chance baseline,
both as fractions. CORE-Extended applies the same rule to the larger
task set specified by Parcae.

MATH uses four demonstration problems, while GSM8K uses a zero-shot
question-answer prompt. Greedy generation selects the highest-probability
next token at every step, with a 256-token cap.
GSM8K extracts the final numerical answer and compares it with the
reference; MATH verifies mathematical equivalence symbolically.
For zero-shot MathQA, answer choices are ranked by length-normalized
log likelihood and accuracy is the fraction of correct selections.
MMLU College Mathematics and Abstract Algebra each contain 100 test
questions. The adaptation panel also includes MMLU High School
Mathematics and WinoGrande, which tests contextual commonsense reasoning
with 1,267 zero-shot questions. BB-Math denotes BIG-Bench Elementary Math QA.
Arithmetic is the \texttt{simple\_arithmetic\_withspaces} task, which
tests completion of arithmetic expressions.
PubMedQA uses the labeled set and LSAT-AR uses AGIEval's analytical
reasoning task. All reported scores are percentages or percentage-point
aggregates, with larger values indicating better performance.
Loop Transformer and \mota\ generation use the trained recurrent forward
at eight loops. The 140M \mota\ and 770M recurrent generation records use
full-prefix evaluation, processing the available token sequence together.
The remaining records use the vLLM GPU inference engine. Initial states
are determined by the token identity, absolute position, channel, and
random seed, with separate key-value caches for each visit to the
shared layers.

\section{Detailed Comparison with Related Architectures}
\label{app:comparison}

\mota\ makes the result of one attention read specify a desired property
of the next read. Its contribution is an explicit relation between
\emph{what was selected}, \emph{what should be selected next}, and
\emph{how the current sources can realize that request}. This section
positions that relation within recurrent computation, attention design,
adaptive memory, and optimization-based architectures.
\Cref{tab:architecture_comparison} compares the information each method
carries and the operation it changes. The following analysis develops
the corresponding mathematical distinctions.

\begin{table}[!htbp]
 \centering
 \caption{\textbf{Architectural comparison of recurrent computation and
 attention updates.} The comparison follows the quantity each method
 updates and the operation that uses it. The last column identifies
 the corresponding design choice in \mota.
 DCMHA abbreviates Dynamically Composable Multi-Head Attention;
 TTT abbreviates test-time training.}
 \label{tab:architecture_comparison}
 \begingroup
 \small
 \setlength{\tabcolsep}{6pt}
 \renewcommand{\arraystretch}{1.16}
 \begin{tabularx}{\linewidth}{@{}>{\raggedright\arraybackslash}p{.22\linewidth}
                            >{\raggedright\arraybackslash}X
                            >{\raggedright\arraybackslash}X@{}}
 \toprule
 Architecture & State and defining operation & \mota's design contribution\\
 \midrule
 Universal Transformer; recurrent-depth language models; Parcae
 \cite{dehghani2019universal,geiping2025recurrent,prairie2026parcae}
 & Repeatedly transform hidden states with a shared core; recompute
 attention from the evolving representations.
 & Give a read an explicit next-read target and compare it with
 the current source-coordinate mean.\\
 \addlinespace[3pt]
 Feedback Transformer \cite{fan2020feedback}
 & Make higher-level representations of previous tokens available
 to future token computations.
 & Carry a target between core visits to direct source selection
 within the next attention operation.\\
 \addlinespace[3pt]
 RealFormer \cite{he2021realformer}
 & Propagate pre-softmax attention scores through a residual path.
 & Compute the score correction from a target discrepancy and
 the statistics of the current sources.\\
 \addlinespace[3pt]
 Attention Residuals \cite{kimi2026attentionresiduals}
 & Attend over earlier layer or block outputs to form the
 next depth representation.
 & Use earlier computation to change attention over causal
 token sources.\\
 \addlinespace[3pt]
 Compositional Attention; DCMHA
 \cite{mittal2022compositional,xiao2024dcmha}
 & Dynamically pair search and retrieval mechanisms, or compose
 scores and weights across heads.
 & Learn how the sources selected on one visit specify a
 target for the next visit.\\
 \addlinespace[3pt]
 Differential Transformer \cite{ye2025differential}
 & Subtract two softmax maps to cancel shared attention noise.
 & Exponentially reweight a distribution toward a moment target;
 use the weights for both content and the next target.\\
 \addlinespace[3pt]
 Gated DeltaNet; Gated DeltaNet-2
 \cite{yang2025gateddelta,hatamizadeh2026gateddelta2}
 & Edit a compressed associative memory through decay,
 erasure, and delta-rule writing.
 & Correct a reading discrepancy by redistributing attention
 over the explicit causal prefix.\\
 \addlinespace[3pt]
 TTT; Titans \cite{sun2025ttt,behrouz2025titans}
 & Update predictor or neural-memory parameters online
 to encode token history.
 & Optimize query-specific tilt coefficients using current
 source moments and a previous reading target.\\
 \addlinespace[3pt]
 Modern Hopfield networks; Energy Transformer
 \cite{ramsauer2021hopfield,hoover2023energy}
 & Formulate associative retrieval or token-state dynamics
 through an energy function.
 & Define an attention-distribution objective conditioned
 on the preceding read, with a source-dependent analytic step.\\
 \addlinespace[3pt]
 Entmax \cite{peters2019entmax}
 & Change the score-to-distribution regularizer to produce
 sparse attention.
 & Match a desired source mean while preserving relevance
 through a KL penalty to base attention.\\
 \addlinespace[3pt]
 \oursrow
 \textbf{\mota}
 & \textbf{Carry a learned moment target; compute a covariance-based
 exponential tilt with a source-range step.}
 & \textbf{The same distribution reads content and writes the
 next target, closing the relation between successive reads.}\\
 \bottomrule
 \end{tabularx}
 \endgroup
\end{table}

\subsection{From Recurrent Computation to a Recurrent Reading Objective}
\label{app:comparison_recurrence}

Universal Transformers apply a shared transformation repeatedly across
depth \cite{dehghani2019universal}. Recurrent-depth language models develop
this idea into repeated latent computation \cite{geiping2025recurrent},
and Parcae studies its scaling with model size and training data
\cite{prairie2026parcae}. At the attention level, the evolving hidden state
produces new queries, keys, and values on each visit. Thus, a loop already
changes which information is read. The additional opportunity is to give
the result of a read an explicit role in specifying the next selection.

Suppressing token and head indices, this distinction can be expressed as
\begin{align}
 \text{Loop Transformer:}\quad
 &p^t=\softmax(\ell(H^t)),\qquad
   y^t=\E_{p^t}[v^t],\qquad H^{t+1}=\mathcal F_\theta(H^t,y^t;x),
 \label{eq:comparison_loop}\\
 \text{\mota:}\quad
 &a^t=\mathcal A(p^t,n^t,m^{t-1}),\qquad
   (y^t,m^t)=\big(\E_{a^t}[v^t],\E_{a^t}[w^t]\big).
 \label{eq:comparison_mota}
\end{align}
Here $\ell(H^t)$ denotes query-key scores computed from state $H^t$,
$x$ is the input, and $\mathcal F_\theta$ includes the remaining shared-core
computation and input injection with learned parameters $\theta$.
\mota\ supplies its content output to the same
computation, while $\mathcal A$ is the moment update in
\Cref{sec:method}. The new state $m^{t-1}$ has a defined meaning: it is
the desired mean of the current read coordinates. The difference
$m^{t-1}-\E_{p^t}[n^t]$ measures how far the default selection is from
that request. This gives the recurrent architecture a direct operation
for revising information access.

The separate write projection is central to this relation. A source
can be selected according to one property, represented by $n_j$, and
request a subsequent property through $w_j$. Training learns this
association through the language-model objective. For a sequence of
dependent lookups, it provides a route by which the information selected
now determines the information needed afterward. The full content value
still enters the hidden-state computation, so the target and the content
serve distinct, coordinated purposes.

Feedback Transformer exposes higher-level representations of previous
tokens to subsequent token computations \cite{fan2020feedback}. Its
feedback enriches the representations available as memory. \mota\ carries
a reading target between visits to the shared core and compares that
target with the current selection. This places the feedback at the
decision about which causal sources to read. The recurrent state therefore
specifies an operation on available knowledge, and the next read writes
the successor of that operation.

\subsection{Attention Scores, Head Composition, and Depth Selection}
\label{app:comparison_attention}

\paragraph{Accumulating scores and correcting a reading discrepancy.}
RealFormer propagates pre-softmax scores through a residual attention
path \cite{he2021realformer}. In its running-sum form, layer $l$ uses
\begin{equation}
 R^l=Q^l(K^l)^\top/\sqrt{d_h}+R^{l-1},\qquad
 A^l=\softmax(R^l).
 \label{eq:comparison_realformer}
\end{equation}
Here $R^l$ stores accumulated logits and $A^l$ is the attention matrix
at layer $l$. The carried quantity records accumulated source preferences. \mota\
carries a target in learned source coordinates and computes a new score
correction from the sources at the current visit. In particular,
when $m=\mu$, its correction is exactly zero; changing the current
covariance changes the correction even if $m-\mu$ stays fixed.
These properties follow from $d=(\Sigma+\lambda I)^{-1}(m-\mu)$.
They make the update responsive to both the reading request and the
source set that must satisfy it.

\paragraph{Denoising and target-directed reweighting.}
Differential Transformer subtracts two attention maps to suppress shared
noise \cite{ye2025differential}. Before its output normalization, a head
has the form
\begin{equation}
 Y_{\mathrm{diff}}=
 \big[\softmax(L_1)-\gamma\softmax(L_2)\big]V.
 \label{eq:comparison_differential}
\end{equation}
Here $L_1,L_2$ are two query-key score matrices, $\gamma$ controls their
subtraction, and $V$ contains the values. This creates signed weights
through a difference of distributions.
\mota\ changes one probability distribution by exponential reweighting,
$a_j\propto p_j\exp(\alpha^\top n_j)$. Its direction is determined by the
target discrepancy, and its step is determined by the range of source
score changes. The resulting weights simultaneously retrieve content
and generate the next target. Denoising and sequentially directing a
read therefore give different mathematical roles to the attention update.

\paragraph{Composing heads and choosing depth representations.}
Compositional Attention separates query-key search from value retrieval
and dynamically pairs their mechanisms \cite{mittal2022compositional}.
Dynamically Composable Multi-Head Attention (DCMHA) transforms score and weight
matrices across heads in an input-dependent manner
\cite{xiao2024dcmha}. Both expand how attention heads combine operations
within a layer. \mota\ instead learns a relation across successive reads:
the same selected sources provide content now and a target for the next
visit. Separate read and write coordinates implement this temporal
relation without requiring a change to the head-composition rule.

Attention Residuals selects among preceding layer or block outputs
\cite{kimi2026attentionresiduals}. Its attention axis is depth, and the
selected values are earlier representations of the same token.
\mota\ selects causal token sources within the recurrent attention
layer. Earlier computation enters as a target that changes those token
weights. Depth selection determines which representations to combine;
moment-directed selection determines which information the next
representation should be built from.

\subsection{Updating a Memory and Directing Its Use}
\label{app:comparison_memory}

Linear attention compresses token history into a recurrent state
\cite{katharopoulos2020linear}, and fast-weight formulations interpret
this state as an associative memory \cite{schlag2021fastweights}.
Gated DeltaNet combines forgetting with a delta-rule edit, which corrects
the difference between a stored prediction and the desired value
\cite{yang2025gateddelta}. Using a value-by-key matrix $S_s$ and token
index $s$, its update can be written as
\begin{equation}
 \bar S_s=\gamma_sS_{s-1},\qquad
 S_s=\bar S_s+\beta_s(v_s-\bar S_sk_s)k_s^\top,\qquad
 o_s=S_sq_s.
 \label{eq:comparison_delta}
\end{equation}
The matrix $S_s$ maps keys to values, $\gamma_s$ controls forgetting,
$\beta_s$ controls write strength, and $o_s$ is the output for query $q_s$.
The residual $v_s-\bar S_sk_s$ measures an association that the memory
should correct. Gated DeltaNet-2 further separates channel-wise erase
and write decisions while retaining channel-wise decay
\cite{hatamizadeh2026gateddelta2}. These rules control what is stored in
a compressed memory as tokens arrive.

\mota\ uses a residual with a different object and effect:
$m-\E_p[n]$ asks the current attention distribution to change which
sources it reads. The update acts on selection probabilities over the
explicit causal prefix. Its $r$-dimensional state stores a desired
source property, while the original values retain the content to be
retrieved. This separates the amount of state needed to direct a read
from the dimension of the knowledge representation that the read returns.

Test-time training (TTT) layers use the parameters of a predictor as their
recurrent hidden state and update them with a self-supervised loss
\cite{sun2025ttt}:
\begin{equation}
 \Theta_s=\Theta_{s-1}-\epsilon_s
     \nabla_\Theta\mathcal L_s(\Theta_{s-1}),\qquad
 o_s=f_{\Theta_s}(q_s).
 \label{eq:comparison_ttt}
\end{equation}
Here $\Theta_s$ contains the predictor's parameters, $\epsilon_s$ is its
update step size, $\mathcal L_s$ is the self-supervised loss, and
$f_{\Theta_s}$ maps query $q_s$ to output $o_s$.
Titans develops an online neural memory updated by gradients of its
prediction loss, with momentum and forgetting, and connects it to attention
\cite{behrouz2025titans}. These architectures learn a memory update
whose resulting parameters encode past information. The inner
optimization in \mota\ acts on $r$ tilt coefficients for the current
query. Its gradient is a reading discrepancy, its Hessian is the
covariance of current source coordinates plus a ridge, and its output
is a revised distribution. The learned projections define the coordinate
system; a small analytic solve determines how the current state uses it.
This makes optimization a component of the attention read itself.

\subsection{From an Attention Energy to a Source-Conditioned Update}
\label{app:comparison_geometry}

Modern Hopfield networks connect attention to associative retrieval and
an energy-based state update \cite{ramsauer2021hopfield}. Energy
Transformer constructs token interactions through descent on a designed
energy function \cite{hoover2023energy}. \mota\ gives the attention
distribution an objective tied to the preceding read:
\begin{equation}
 \min_{a\in\Delta}
   \lambda\KL(a\Vert p)+\tfrac12\norm{\E_a[n]-m}^2.
 \label{eq:comparison_objective}
\end{equation}
The base distribution $p$ supplies the current relevance structure,
the target $m$ supplies the desired source property, and the coordinates
$n$ specify the properties available for selection. This objective
turns cross-loop state into a concrete operation on attention.

Entmax changes the regularizer used to map scores to an attention
distribution, allowing some source probabilities to be exactly zero
\cite{peters2019entmax}.
In \Cref{eq:comparison_objective}, the additional term depends on the
attention-weighted mean $\E_a[n]$: changing one source's probability affects
the same target residual as changing any other. The construction thus
specifies which source properties the distribution should realize.
Its dual exposes the dependence directly:
\begin{equation}
 \underbrace{\nabla G(0)}_{\text{reading discrepancy}}=\mu-m,
 \qquad
 \underbrace{\nabla^2G(0)}_{\text{source geometry}}=\Sigma+\lambda I,
 \qquad
 \underbrace{\alpha}_{\text{score correction}}
 =\frac{\log(1+\rho)}{\rho}(\Sigma+\lambda I)^{-1}(m-\mu).
 \label{eq:comparison_geometry}
\end{equation}
The covariance is also the Fisher information of the exponential tilt
family, connecting this construction to natural-gradient geometry
\cite{amari1998natural,martens2020natural}. Here that geometry is
computed from each query's available sources and used inside its
forward attention operation.

The source range $\rho$ makes the finite step follow the same geometry
as the objective. A target component shared equally by every source
cannot change their relative probabilities. The complete \mota\ update
therefore ignores such a component, including when determining the
step size (\Cref{prop:invariance}). A norm-based step can shrink the
useful correction in response to that component. By measuring the
range of $n_j^\top d$, \mota\ sets the step using exactly the distinctions
that affect selection. The resulting rule both decreases the per-read
dual objective and bounds the change in pairwise attention odds
(\Cref{thm:descent}). These results explain why the direction and the
step belong to one design.

\subsection{What the Low-Dimensional State Contributes}
\label{app:comparison_contribution}

The added logit matrix has the factorized form
$\Delta L=B N^\top$, where row $i$ of $B$ is $\alpha_i^\top$
and row $j$ of $N$ is $n_j^\top$. This gives the correction rank at most
$r$ before causal masking and permits the augmented query-key
implementation in \Cref{eq:augmented}. The distinguishing choice is
how those factors are obtained. A directly parameterized correction
would predict $\alpha$ with a learned mapping. \mota\ computes it by
comparing a previous target with the current source mean, solving with
the current covariance, and taking a source-range step. The factorization
provides the implementation; the reading objective determines its behavior.

This behavior can be followed all the way to the output. At $m=\mu$,
\Cref{prop:response} gives the Jacobian $D_m$, the derivatives of the
content and next target with respect to the incoming target:
\begin{equation}
 D_m
 \begin{bmatrix}y\\m_{\mathrm{next}}\end{bmatrix}
 =
 \begin{bmatrix}\Cov_p(v,n)\\\Cov_p(w,n)\end{bmatrix}
 (\Sigma+\lambda I)^{-1}.
 \label{eq:comparison_response}
\end{equation}
The regularized inverse covariance converts the requested change into source
reweighting. The upper cross-covariance converts that reweighting into
retrieved content; the lower one converts it into the next target.
Separate projections let the two outputs express different properties
of the same selected information. A compact target can consequently
direct a full-dimensional content read and establish a new reading
request through one shared distribution.

The architectural contribution consists of three connected choices:
representing the next read by a learned moment target, realizing that
target through the geometry of the current sources, and making the
resulting read write its successor. Each choice has a mathematical role
and a corresponding analysis. The dimension and placement ablations
examine where the target operates (\Cref{tab:dimension,tab:placement});
the covariance and mean-subtraction ablations examine how the request becomes
a distribution (\Cref{tab:geometry}); the source-range analysis explains
the finite step (\Cref{thm:descent}). The feedback and loop-depth
interventions examine how the trained model uses successive reads
(\Cref{tab:feedback,tab:depth}). Together they connect
\mota's motivation, update rule, and experimental analysis through
the same mechanism: selected information actively directs subsequent
information access.

\clearpage
\section{Case Studies: Reading the Conditions Needed Next}
\label{app:attentioncases}
\setlength{\intextsep}{4pt plus 1pt minus 1pt}
\captionsetup[figure]{skip=3pt}

The motivation in \Cref{fig:motivation} concerns the connection between
successive reads: a useful intermediate result changes which information
should be retrieved next. We examine this connection through nine selected
MATH examples, relating saved solutions to attention over the conditions
in each problem. The examples cover inverse-function composition,
quantity conversion, combinatorial selection, algebraic relations,
expected value, and coordinate geometry. Together they show how the
next-read question arises in different mathematical computations.

\paragraph{Paired observations at a fixed reasoning position.}
We use the 1.3B Loop Transformer and \mota\ checkpoints after 20,693
FineMath updates, corresponding to 10.85B mathematical tokens.
For each example, both models receive the original four-shot evaluation
prompt followed by the same explicitly supplied reasoning prefix.
We observe the first attention layer of the shared core over all eight
loops, averaging all twelve heads. One view uses the final query token;
the other averages the final four query tokens. The recorded answers
shown above each figure come from the original free-generation evaluation.
The paired prefixes align the computation being inspected so that a
difference in the generated wording does not change the query context.
We mark the required and comparison source spans from the problem
semantics before extracting their attention. Eight query positions are
selected for clear final-loop source preference in both views; the
original inverse-function example retains both lookup positions.

\paragraph{What the visualizations measure.}
Let $\bar\pi_{tj}^{M}$ be the attention to source token $j$ at loop $t$,
averaged over the selected queries and heads of model $M$.
For a source span $S$, its total attention is
$A_t^M(S)=\sum_{j\in S}\bar\pi_{tj}^M$.
Let $P$ contain the current problem's source tokens. Heatmap cells show
the within-problem attention density
\begin{equation}
 D_t^M(S)=100\,\frac{A_t^M(S)}{|S|\,A_t^M(P)}.
 \label{eq:caseheat}
\end{equation}
Dividing by the token count $|S|$ makes conditions of different lengths
comparable. For a required span $S$ and comparison span $C$, the curves
show
\begin{equation}
 R_t^M(S,C)=
 \frac{A_t^M(S)/|S|}{A_t^M(C)/|C|},
 \qquad
 G_8=\frac{R_8^{\mota}(S,C)}{R_8^{\mathrm{Loop}}(S,C)}.
 \label{eq:casepreference}
\end{equation}
The horizontal line at one indicates equal attention density; the
annotated multiplier is $G_8$. A comparison condition can become useful
at another reasoning step, as the two prices do in a budget calculation.
Heatmaps display loops 1, 2, 4, and 8, while the curves retain every loop.

\paragraph{Connecting source preference to the method.}
In \Cref{eq:jointread}, an attention read returns content and writes a
learned target. At the next visit, the residual $m-\mu$ is interpreted
through $(\Sigma+\lambda I)^{-1}$ and the source-score range
(\Cref{eq:direction,eq:step}). The resulting correction changes relative
source weights according to \Cref{eq:odds}:
\begin{equation}
 \log\frac{a_j/a_k}{p_j/p_k}
   =\alpha^\top(n_j-n_k).
 \label{eq:caserelative}
\end{equation}
The marked conditions provide a readable view of this source allocation;
the learned target itself is a vector of source features rather than a
literal storage of a numerical intermediate answer. The figures compare
the final attention of two trained models. Within \mota, the base
distribution $p$ and corrected distribution $a$ provide a second,
local comparison. For example, the price query has final-loop ratios
2.137 for Loop Transformer, 4.201 for \mota's base read, and 4.257 after
its correction. The marked span's absolute mass can decrease while its
ratio to the comparison span increases, because attention is normalized
over the entire prefix. This distinguishes the learned recurrent
representation from the additional redistribution made on one visit.

\clearpage
\subsection{An Intermediate Result Changes the Next Lookup}
\label{app:caseinverse}

\begin{figure}[H]
 \centering
 \includegraphics[width=\linewidth,trim=0pt 11pt 0pt 11pt,clip]{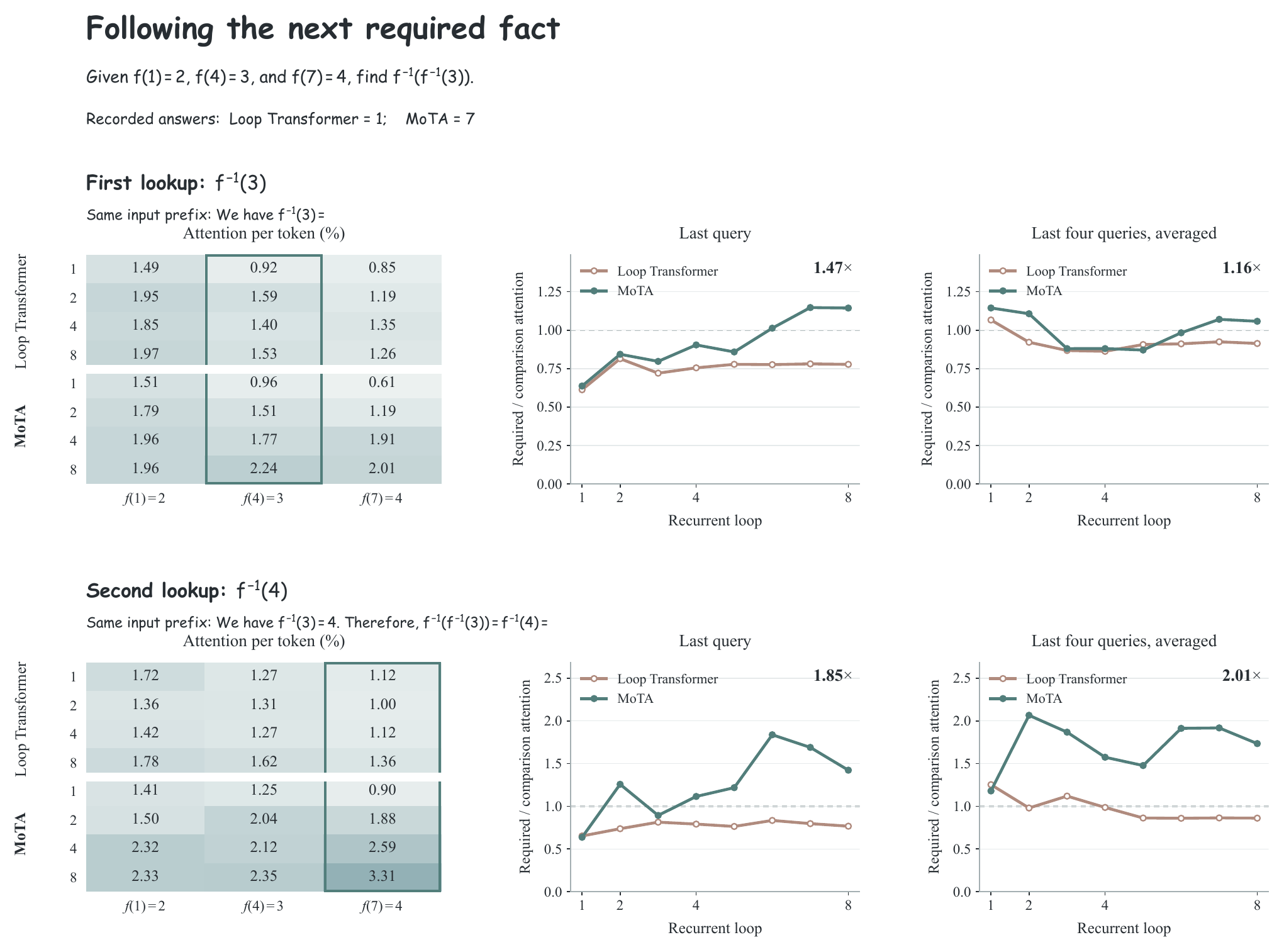}
 \caption{\textbf{Inverse-function composition.} The first query requires
 $f(4)=3$; after obtaining the intermediate result $4$, the second requires
 $f(7)=4$. The outlined source changes accordingly. Both queries use
 $f(1)=2$ as the comparison condition.}
 \label{fig:caseinverse}
\end{figure}

The required sequence is $3\mapsto4\mapsto7$. The saved baseline
solution initially lists the correct inverse mappings, then contradicts
its own $f^{-1}(3)=4$ and returns 1. \mota\ carries the intermediate
result into the second lookup and returns 7.
At the matched prefixes in \Cref{fig:caseinverse}, the final-query
preference ratios at loop eight are 0.777 versus 1.144 for the first
lookup and 0.767 versus 1.423 for the second. Thus, \mota\ favours the
currently required fact over the comparison fact at both positions,
and the final-four-query view preserves this ordering.
This example directly instantiates the motivation: information obtained
from one relation determines which different relation is needed next.
The read/write separation in \Cref{sec:readwrite} gives the shared
attention layer an explicit learned state for organizing that transition.

\clearpage
\subsection{Preserving a Quantity Across a Change of Context}
\label{app:casequantity}

\begin{figure}[H]
 \centering
 \includegraphics[width=\linewidth,trim=0pt 11pt 0pt 11pt,clip]{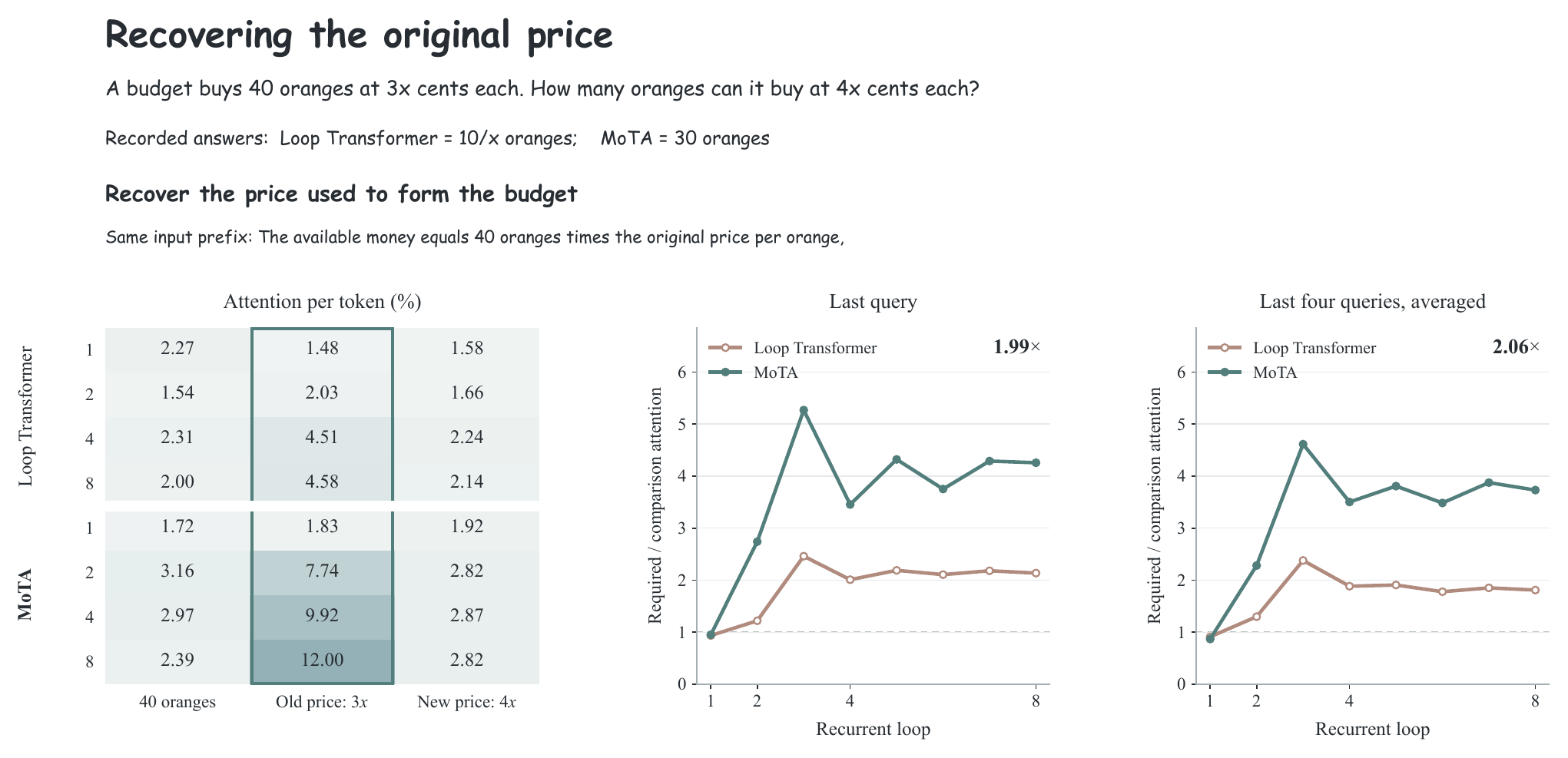}
 \caption{\textbf{Recovering the original price.} The fixed prefix is
 forming the budget, so the required source is the old price $3x$;
 the new price $4x$ is used in the subsequent division.}
 \label{fig:caseprice}
\end{figure}

The budget is $40\cdot3x=120x$, giving $120x/(4x)=30$ oranges.
The baseline instead divides the original quantity by the new price and
returns $10/x$. At the budget-forming query, \mota's preference for the
old price reaches 4.257, compared with 2.137 for the baseline
(\Cref{fig:caseprice}). The two prices share a semantic role but serve
different steps. Their separation illustrates why a next-read target
should express the current information need rather than generic relevance.

\begin{figure}[H]
 \centering
 \includegraphics[width=\linewidth,trim=0pt 11pt 0pt 11pt,clip]{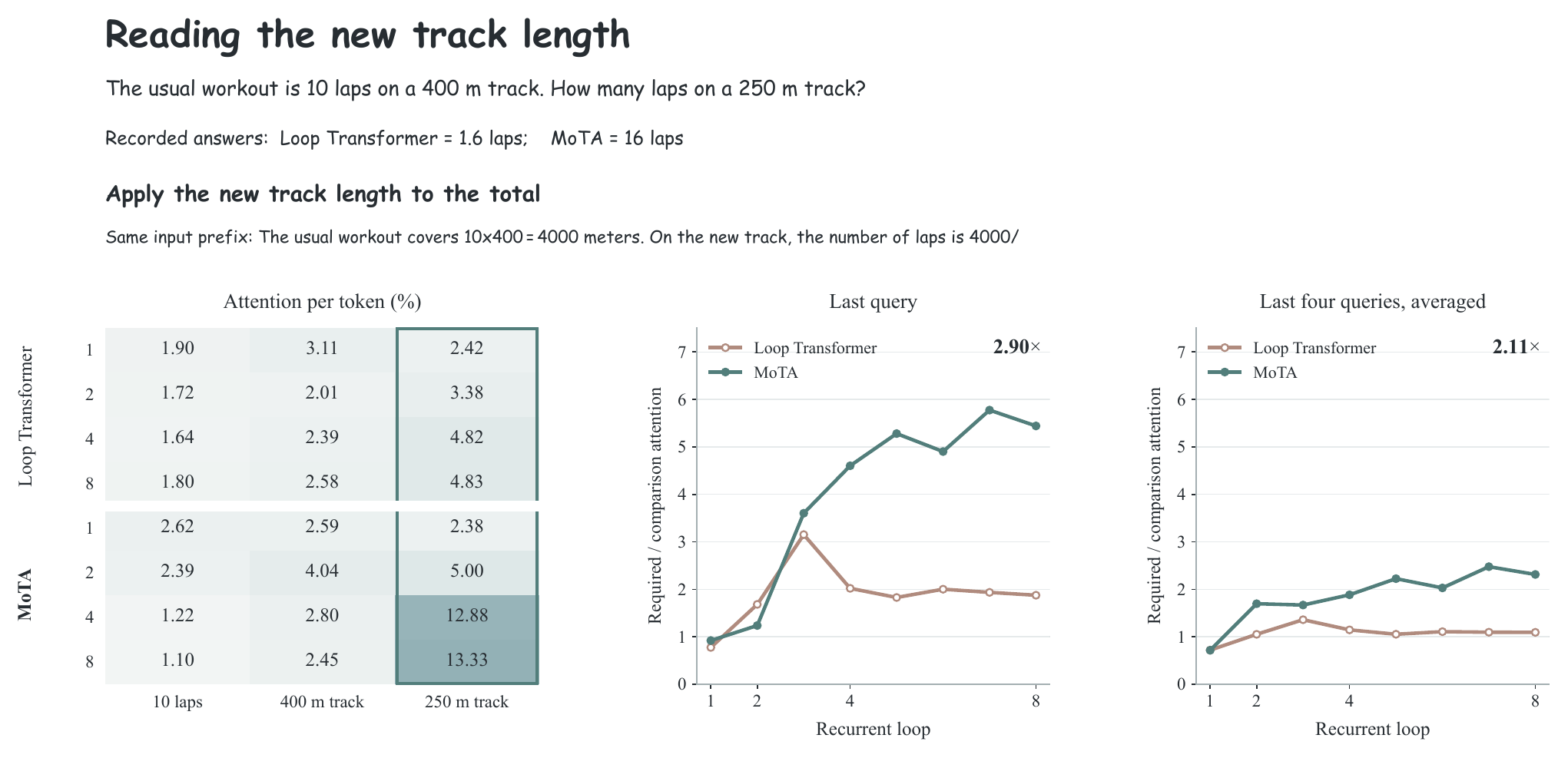}
 \caption{\textbf{Reading the new track length.} After the supplied prefix
 establishes a 4,000-metre workout, the next operation requires the
 250-metre track length rather than the original 400-metre length.}
 \label{fig:casetrack}
\end{figure}

\mota\ computes $10\cdot400/250=16$ laps, whereas the baseline uses
$400/250=1.6$ and omits the original ten-lap quantity.
The plotted prefix already supplies the total distance; at that point,
\mota\ distinguishes the new track length with ratios 5.442 versus
1.876, a 2.90-fold difference (\Cref{fig:casetrack}).
Together, the price and track cases expose the same structure: first
establish a conserved quantity, then retrieve the condition needed to
express it in the new context.

\clearpage
\subsection{Keeping Populations and Selection Requirements Distinct}
\label{app:casegroups}

\begin{figure}[H]
 \centering
 \includegraphics[width=\linewidth,trim=0pt 11pt 0pt 11pt,clip]{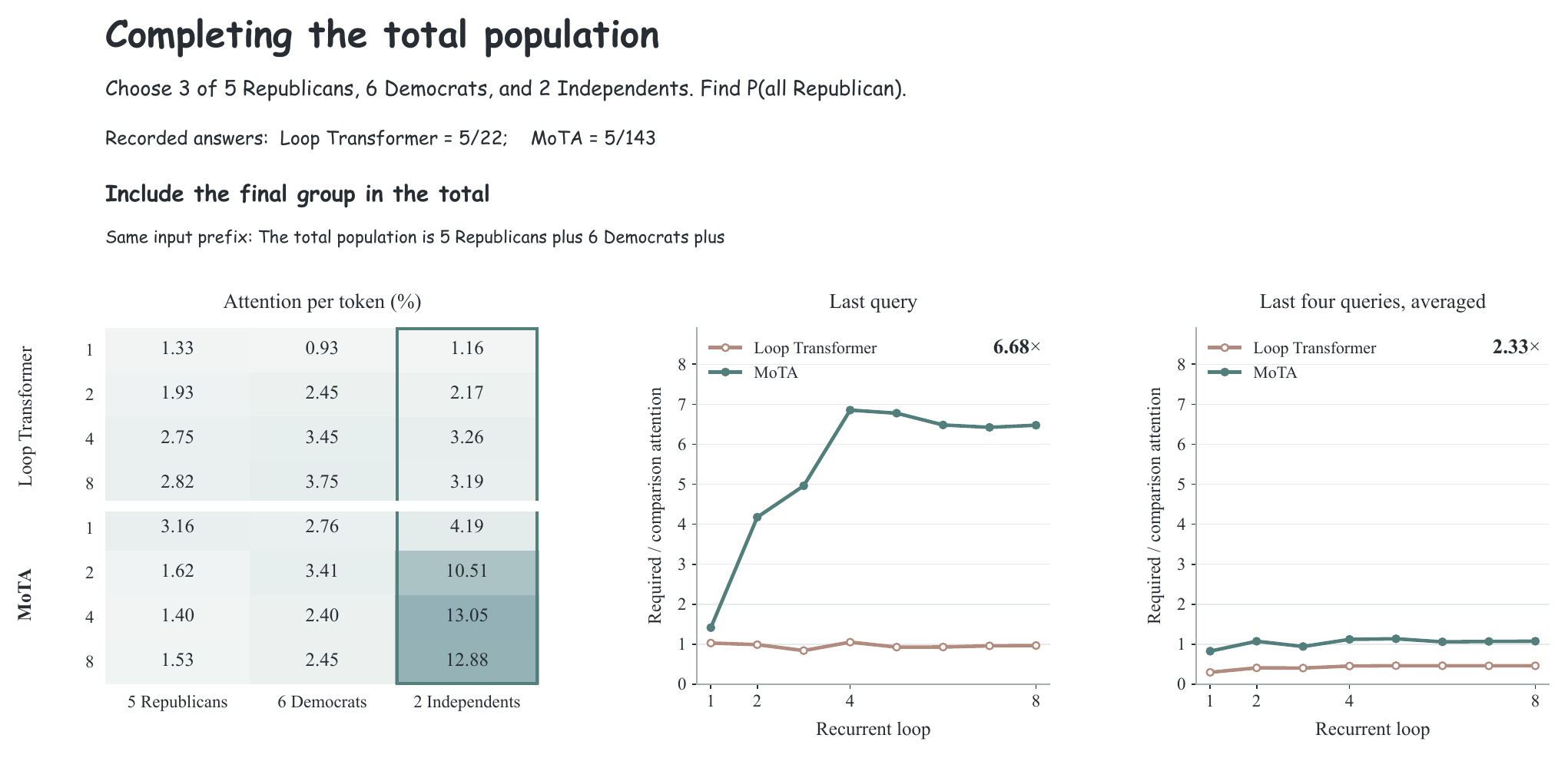}
 \caption{\textbf{Completing the population count.} The supplied prefix
 includes the Republican and Democratic counts; the next required source
 is the two independent members. The other two groups form the comparison.}
 \label{fig:casesenate}
\end{figure}

The correct probability is
$\binom{5}{3}/\binom{13}{3}=5/143$. The baseline uses a population of
12 and makes a further arithmetic error; \mota\ correctly uses all 13
members. At the total-completion query, the final preference is 6.476
for \mota\ and 0.969 for the baseline (\Cref{fig:casesenate}).
The requested source is determined by which counts are already present
in the intermediate computation, matching the next-read principle.

\begin{figure}[H]
 \centering
 \includegraphics[width=\linewidth,trim=0pt 11pt 0pt 11pt,clip]{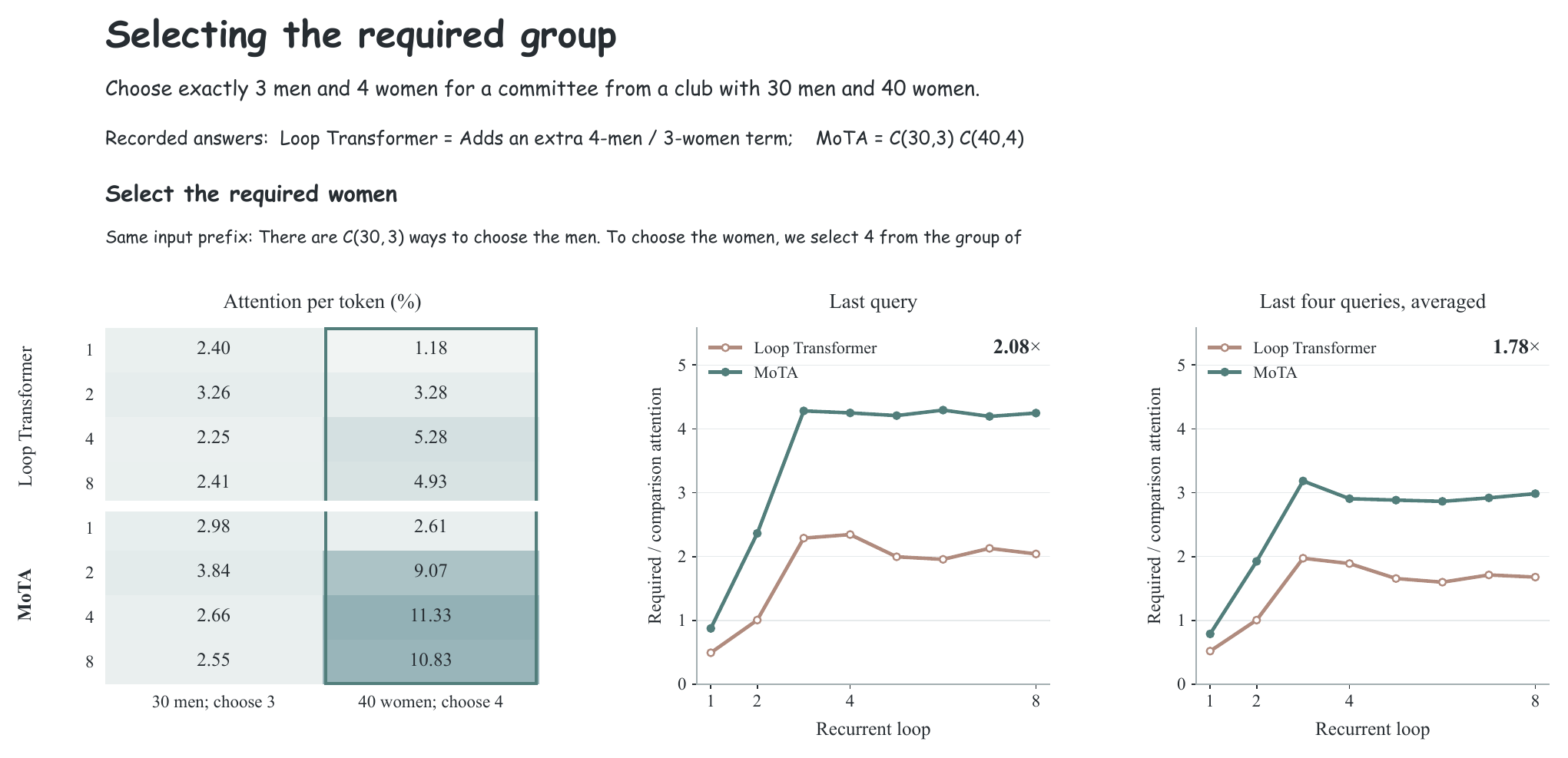}
 \caption{\textbf{Selecting the required female group.} Source spans pair
 each group's population with its requested selection count. After the
 male choices have been formed, the query requires the female group.}
 \label{fig:casecommittee}
\end{figure}

The required count is $\binom{30}{3}\binom{40}{4}=371{,}043{,}400$.
\mota\ returns the equivalent factorial expression, while the baseline
adds an unrequested four-men/three-women composition.
The female-group preference reaches 4.248 versus 2.041
(\Cref{fig:casecommittee}). This case combines a population size and
a selection constraint within the same semantic source group. Such
relations motivate learned source coordinates that organize what a read
should request, rather than a rule based only on token proximity.

\clearpage
\subsection{Selecting the Relation for the Current Algebraic Operation}
\label{app:casealgebra}

\begin{figure}[H]
 \centering
 \includegraphics[width=\linewidth,trim=0pt 11pt 0pt 11pt,clip]{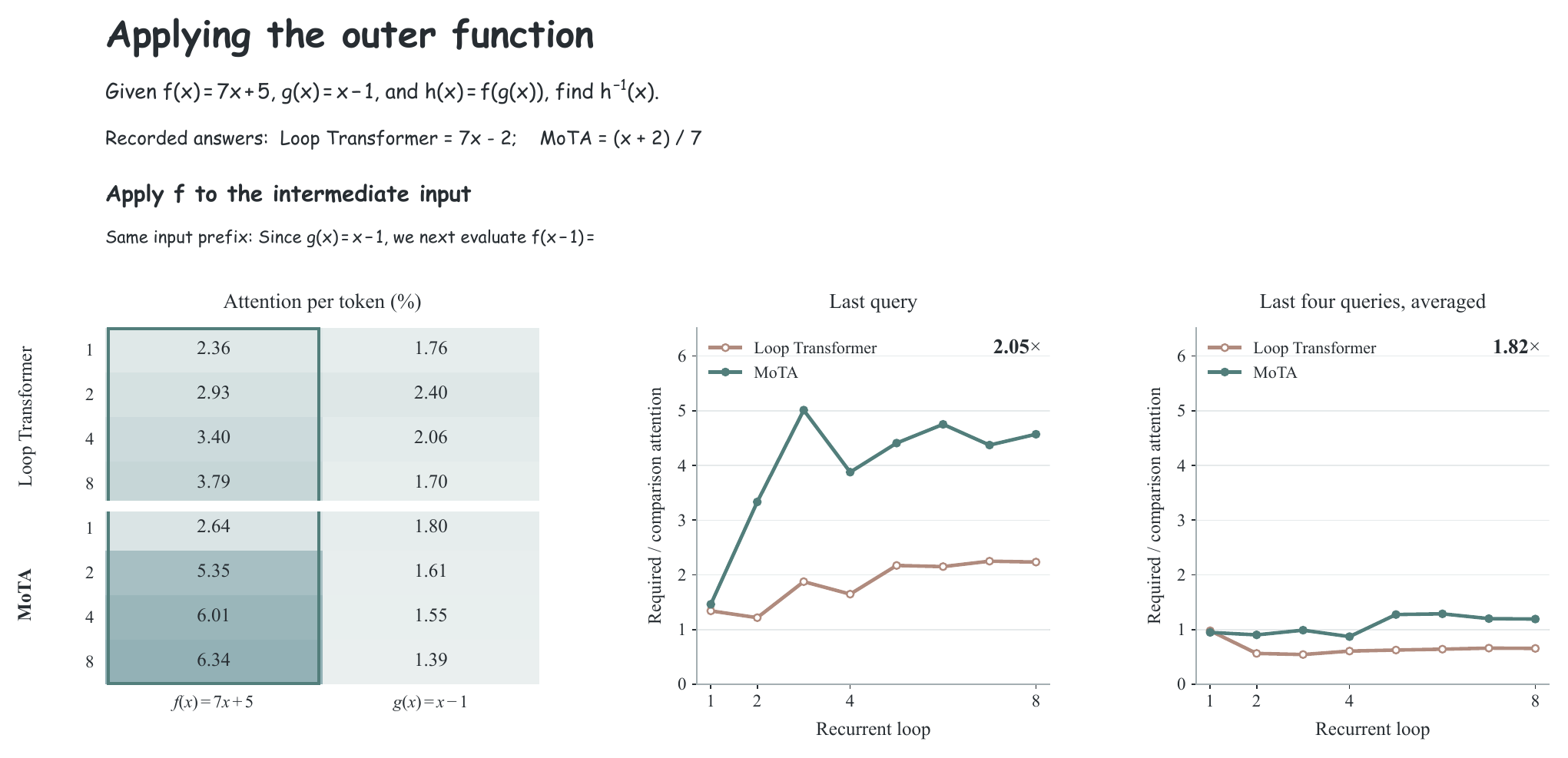}
 \caption{\textbf{Applying the outer function.} Once the supplied prefix
 establishes $g(x)=x-1$, the next substitution requires $f(x)=7x+5$.
 The plotted query concerns composition; the recorded final answers concern inversion.}
 \label{fig:casecompose}
\end{figure}

Both saved solutions correctly form $h(x)=7x-2$. The baseline then
reports this expression as its own inverse, whereas \mota\ returns
$(x+2)/7$. The matched composition query measures how the two trained
models distinguish the outer function from the inner one: the final
ratios are 4.571 and 2.234 (\Cref{fig:casecompose}).
The relevant relation changes after substitution of the intermediate
expression, giving a concrete interpretation to the distinct read and
write roles in \Cref{eq:jointread}.

\begin{figure}[H]
 \centering
 \includegraphics[width=\linewidth,trim=0pt 11pt 0pt 11pt,clip]{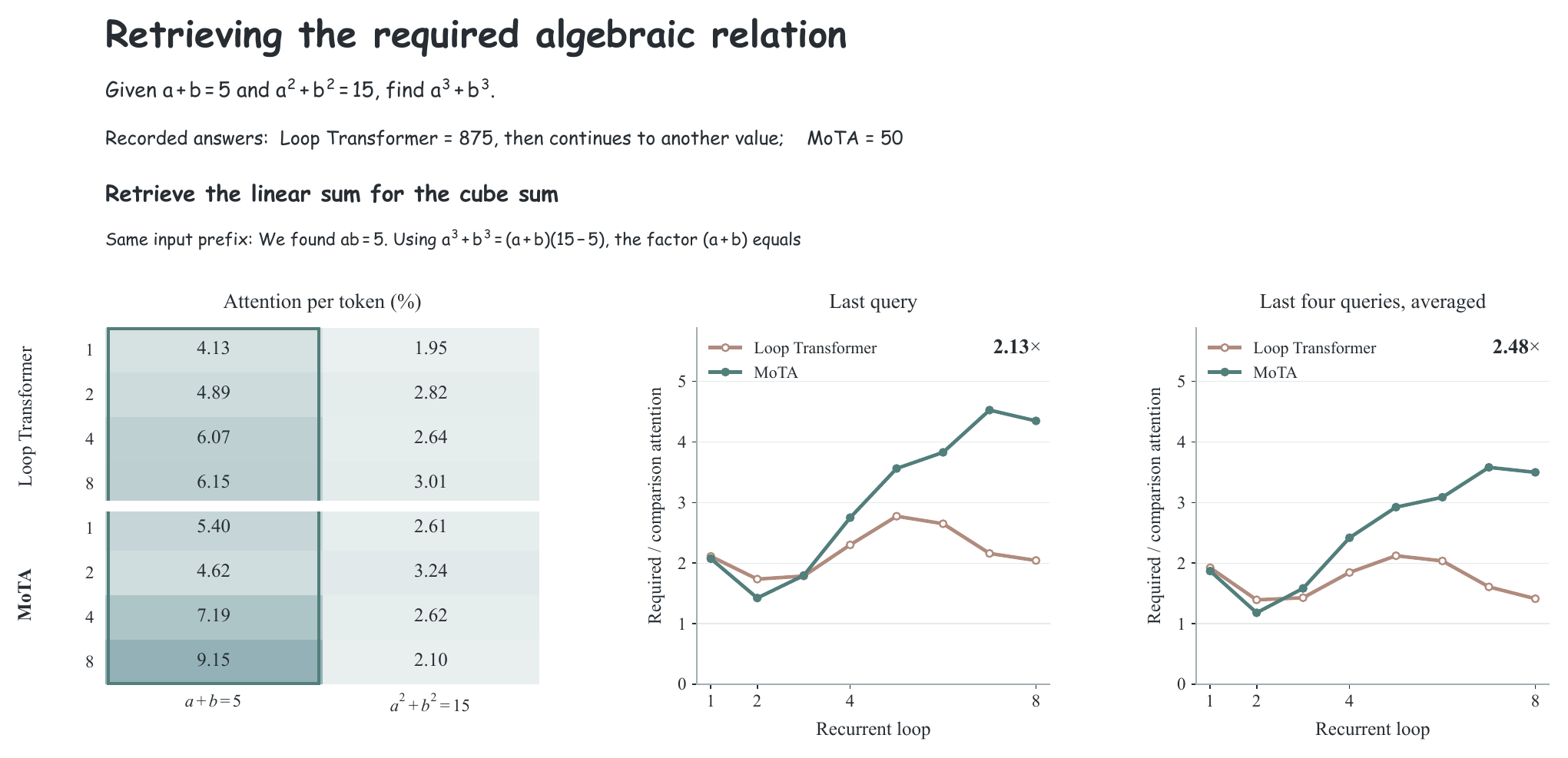}
 \caption{\textbf{Retrieving the linear sum.} Given $ab=5$ in the supplied
 prefix, evaluating $(a+b)(15-5)$ requires the original relation $a+b=5$.
 The square-sum relation is the comparison source.}
 \label{fig:casemoments}
\end{figure}

The relations imply $ab=(25-15)/2=5$, hence
$a^3+b^3=(a+b)(a^2+b^2-ab)=50$.
\mota\ uses this valid identity and returns 50; its response misnames
the identity and leaves the product calculation implicit.
The baseline makes errors in deriving the product and in later algebra.
The plotted final query favours the linear sum with ratios 4.348 versus
2.044 (\Cref{fig:casemoments}), showing how the required source can
change after an intermediate quantity has been established.

\clearpage
\subsection{Reading the Operands of a Local Calculation}
\label{app:caseoperands}

\begin{figure}[H]
 \centering
 \includegraphics[width=\linewidth,trim=0pt 11pt 0pt 11pt,clip]{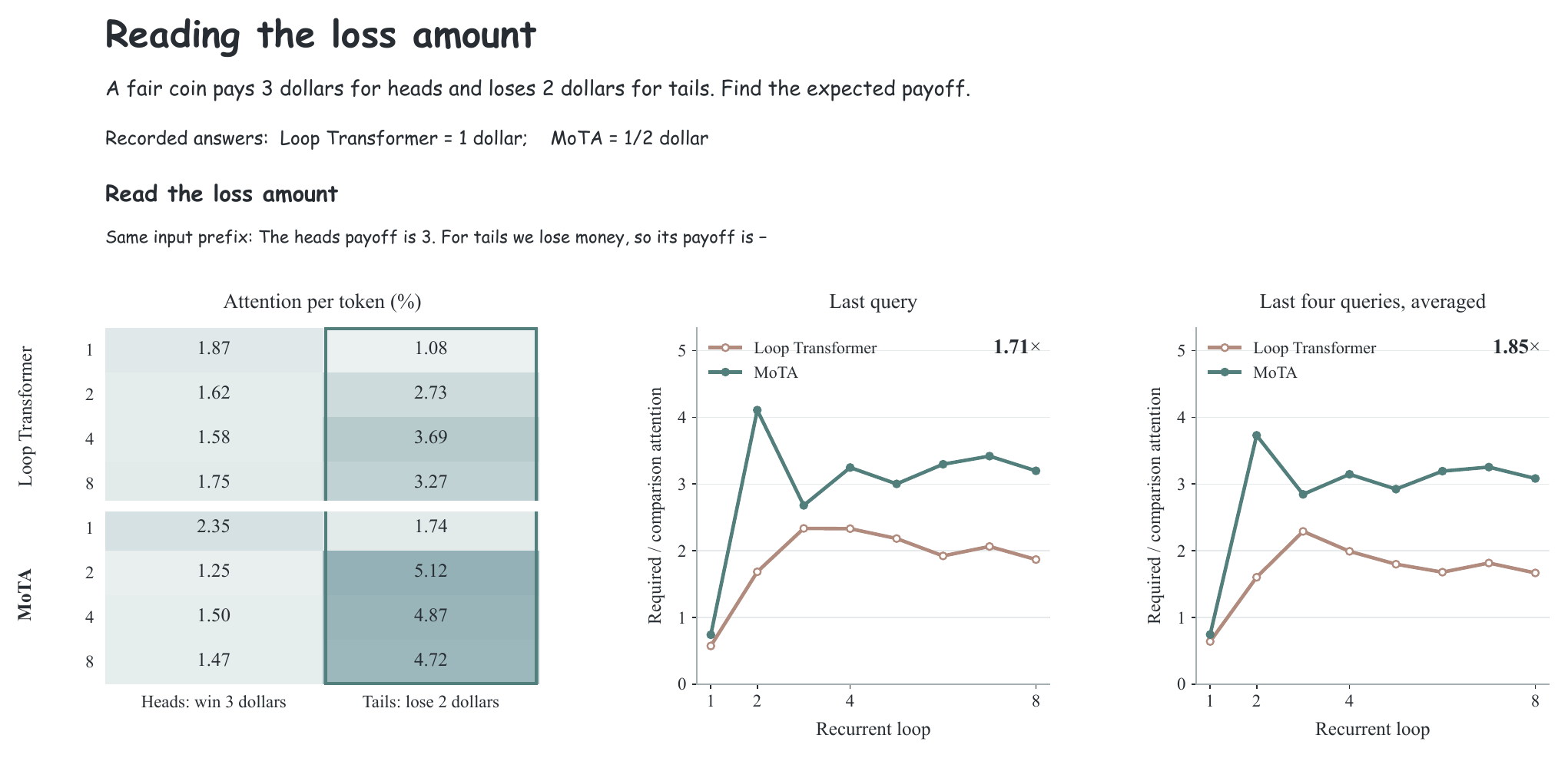}
 \caption{\textbf{Reading the loss amount.} After the heads payoff is
 established, the fixed prefix asks for the magnitude of the negative
 tails payoff. The heads condition is the comparison source.}
 \label{fig:casepayoff}
\end{figure}

The expectation is $\tfrac12(3)+\tfrac12(-2)=\tfrac12$.
Both saved responses set up the expectation correctly; the baseline
then evaluates $1.5-1$ as 1, while \mota\ returns $1/2$.
At the loss-reading query, the source preference is 3.198 versus 1.869
(\Cref{fig:casepayoff}). This example extends the source-allocation
analysis to individual operands: the next read distinguishes the
condition supplying the requested operand from another relevant outcome.

\begin{figure}[H]
 \centering
 \includegraphics[width=\linewidth,trim=0pt 11pt 0pt 11pt,clip]{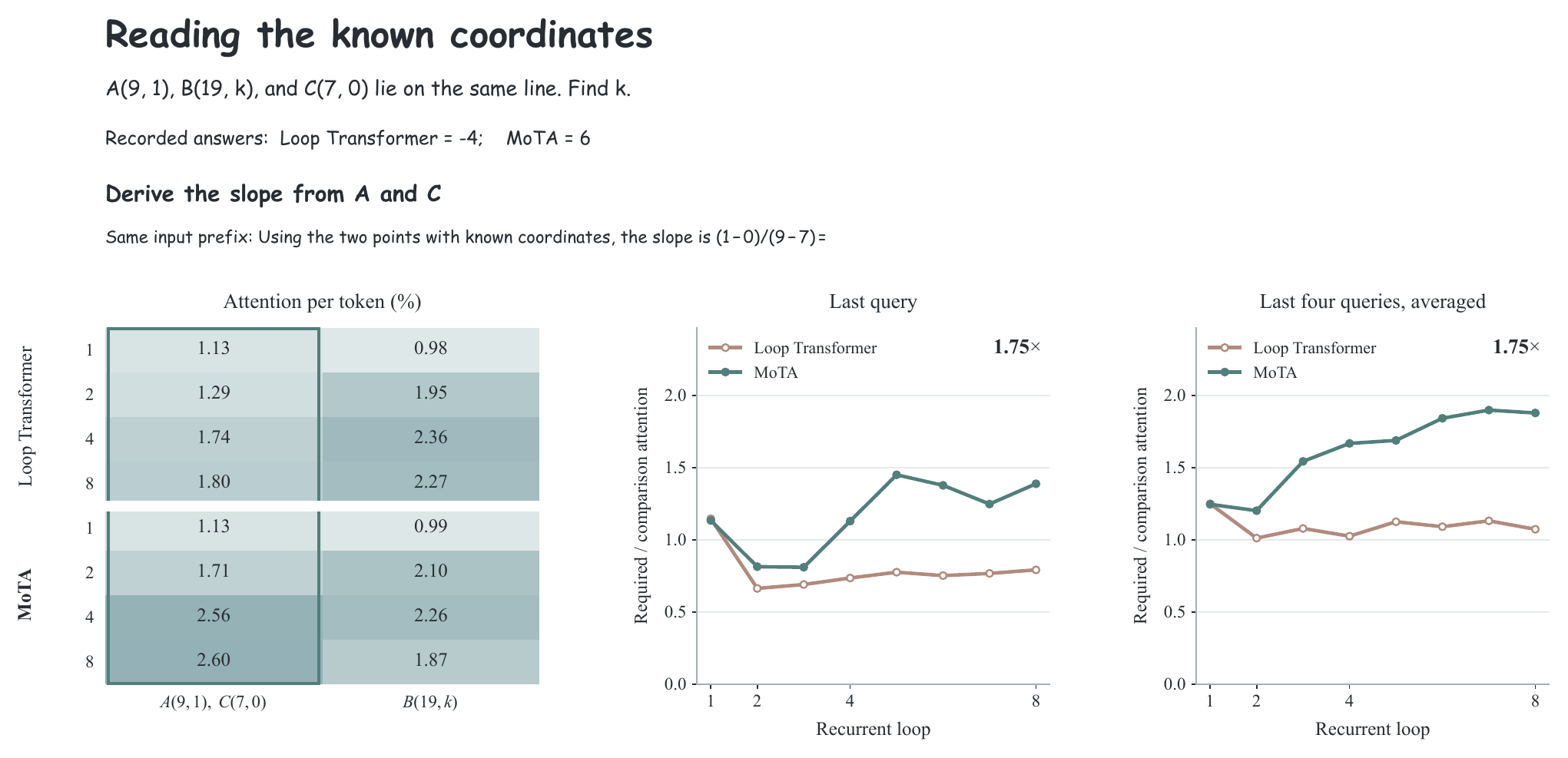}
 \caption{\textbf{Reading the known coordinates.} The slope computation
 requires the two fully specified points $A$ and $C$, compared with
 the point $B$ whose vertical coordinate remains unknown.}
 \label{fig:casepoints}
\end{figure}

The known points give $(1-0)/(9-7)=1/2$, and the position of $B$ then
gives $k=6$. The baseline selects the correct points but evaluates the
slope with the wrong sign; \mota\ computes both steps correctly.
At the fixed slope query, \mota's preference for the known pair is
1.388 versus 0.792 (\Cref{fig:casepoints}). The known pair establishes
the relation subsequently applied to the unknown point, making the
changing reading requirement explicit.

\end{document}